\documentclass[10pt,twocolumn,letterpaper]{article}
\usepackage[pagenumbers]{cvpr} 

\definecolor{cvprblue}{rgb}{0.21,0.49,0.74}
\usepackage{tabularx,array}
\usepackage[pagebackref,breaklinks,colorlinks,allcolors=cvprblue]{hyperref}

\def\paperID{}
\def\confName{}
\def\confYear{2026}
\title{Mechanistic Interpretability for Neural Networks: Circuits, Sparse Features and Symbolic Reasoning}
\author{Pranav Milind Sawant\\
The University of Texas at Dallas\\
800 W Campbell Rd, Richardson, TX 75080, United States\\
{\tt\small pms220001@utdallas.edu}\\
\and
Jakub Krejčí\\
VSB---Technical University of Ostrava\\
708 00, Ostrava, Czech Republic\\
{\tt\small jakub.krejci@vsb.cz}
}
\begin{document}
\maketitle
\begin{abstract}
This article offers a comprehensive overview of mechanistic interpretability, an emerging field that seeks to reverse-engineer the internal algorithms of modern neural networks. While traditional explainable AI methods often stop at surface-level input-output correlations, this approach directly addresses the opaque "black box" nature of machine learning models, which is essential for ensuring safety and auditability in high-stakes deployments. The paper provides a detailed examination of Transformer circuit analysis, exploring how internal components like the residual stream, attention mechanisms, and induction heads drive complex tasks and in-context learning. It subsequently tackles the core challenge of superposition and polysemanticity, demonstrating how tools like Sparse Autoencoders (SAEs) and transcoders can decompose tangled network activations into distinct, human-interpretable features. Furthermore, the paper explores methods for actively controlling and modifying model behavior through steering vectors and causal interventions. Finally, it connects these mechanistic insights with neurosymbolic AI frameworks designed to translate neural representations into explicit, executable logical rules.
\end{abstract}
\section{Introduction}
\label{sec:intro}

Transformer models and large language models (LLMs) have achieved unprecedented success across a wide range of domains; however, their internal decision-making processes remain largely opaque. These systems often function as complex “black boxes”: while we control the input data and observe the final outputs, the computations taking place across hundreds of layers and billions of parameters remain difficult to interpret. This lack of transparency becomes particularly problematic in critical and high-stakes domains such as healthcare, law, or autonomous driving, where trustworthiness, safety, and auditability are essential.

Traditional explainability methods typically focus on correlations between inputs and outputs, for example through saliency maps. Mechanistic interpretability, by contrast, seeks a deeper understanding of a model’s internal computations. Its goal is to reverse-engineer neural networks and transform their learned parameters into human-interpretable algorithms, pseudocode, and functional components \cite{kowalska2025unboxing}. This approach aims to deconstruct models into so-called circuits—smaller subnetworks composed of neurons, attention layers, and other mechanisms that are responsible for specific behaviors.

The importance of mechanistic interpretability is further amplified in the context of AI safety and alignment. Conventional alignment techniques, such as RLHF (Reinforcement Learning from Human Feedback), can train models to exhibit desirable behavior at the surface level, but they do not guarantee that the model has genuinely internalized human values or safe reasoning strategies. Consequently, there is a growing argument that ensuring safe AI requires a shift from purely behavioral control toward understanding the internal mechanisms of these systems \cite{naseem2026mechanisticinterpretabilitylargelanguage}.

At the core of these investigations lies the Transformer architecture \cite{elhage2021mathematical}. Transformers operate using a high-dimensional residual stream, which serves as the model’s central communication channel. Each layer reads from and writes to this stream via linear transformations and additive updates, allowing the model to store different types of information in relatively independent subspaces. Within this structure, attention heads act as mechanisms for information movement. Their behavior can be decomposed into two primary circuits: the Query-Key (QK) circuit determines where to attend in the prior context, while the Output-Value (OV) circuit determines what information to extract and propagate to the current token.

When these operations are mathematically decomposed through path expansion, the Transformer can be interpreted as a large collection of relatively independent computational paths. In their simplest form, these paths resemble skip-trigrams that transfer information from previous tokens based on basic linguistic statistics. However, the true expressive power of Transformers arises from composition, where interactions across layers give rise to more complex circuits. A notable example is induction heads, in which later layers leverage information from earlier layers to detect and complete patterns in the text. By recognizing that a specific token previously followed the current one, the model can accurately predict its recurrence. This mechanism is widely considered a fundamental building block of in-context learning and emergent reasoning capabilities.

A major obstacle to interpreting these systems lies in how neural networks represent information. Rather than assigning each neuron a single well-defined concept (monosemanticity), models frequently employ a strategy known as superposition. This enables them to compress a large number of features into a limited number of dimensions. As a result, neurons become polysemantic: a single neuron may respond to a mixture of seemingly unrelated visual, textual, or abstract concepts. This significantly complicates traditional neuron-level analysis, as meaning is distributed across representations rather than localized within individual units.

Particularly challenging in this regard are the MLP layers, which often behave as opaque “black boxes” within the Transformer architecture itself. The effort to understand their internal representations has led to the development of methods such as Sparse Autoencoders (SAEs), which aim to disentangle polysemantic representations into more interpretable features. Additional tools include Automated Circuit Discovery (ACDC) for systematically identifying internal circuits, and Linear Artificial Tomography (LAT), which enables the analysis and targeted manipulation of internal representations.

This article therefore guides the reader through contemporary approaches in mechanistic interpretability that aim to overcome the opacity of modern neural networks. It begins by introducing the fundamental principles of Transformers, including the residual stream, attention heads, and circuit composition. It then addresses the problem of superposition and polysemanticity as key barriers to interpretability. Subsequently, it explores advanced methods such as ACDC, SAE, and LAT, which enable the analysis, disentanglement, and controlled intervention in model representations. Finally, the article touches on neuro-symbolic artificial intelligence (NeSy), an emerging paradigm that combines the computational power of deep learning with formal logic, enabling the translation of abstract neural representations into precise, executable rules.

\subsection{From Rule based Interpretable ML to Mechanistic Interpretability}
The need for mechanistic interpretability did not emerge in isolation. It developed gradually from earlier attempts to make neural networks more transparent. These earlier methods were primarily designed to answer a practical question: which parts of the input or representation influenced the model’s output? Although they provided useful diagnostic tools, they generally remained focused on surface-level explanations rather than on reconstructing the internal algorithms implemented by the model.

One of the earliest and most widely used families of interpretability methods consists of post-hoc explanation techniques. These methods are applied after a model has already been trained and are used to explain individual decisions made by otherwise opaque machine learning systems. Prominent examples include saliency maps, Grad-CAM, LIME, SHAP, and Guided Backpropagation. Saliency-based methods highlight regions of the input data that had the greatest influence on a model’s decision. For instance, Grad-CAM generates heatmaps using gradient information, while Guided Backpropagation isolates positive signals during backpropagation through the network \cite{kares2025makesgoodsaliencymap}.

Among model-agnostic post-hoc methods, LIME and SHAP are especially influential \cite{Salih_2024}. LIME explains a single prediction by perturbing the input locally and fitting a simplified linear surrogate model around that specific decision. This makes it computationally efficient, but strictly local. SHAP, by contrast, is based on cooperative game theory and estimates feature contributions by analyzing combinations of input features, which allows it to provide both local and global explanations. These methods are valuable because they offer intuitive, human-readable explanations of model behavior.

However, they also have serious limitations. Both LIME and SHAP can be vulnerable to collinearity, because they often rely on assumptions that treat input features as independent, even when real-world data contain strong dependencies between them. Studies also show that their explanations can be highly model-dependent: for the same dataset, classifiers with comparable accuracy may produce substantially different feature-importance rankings. Because these tools can be misled by biased models and because users often place excessive trust in their outputs, additional stability metrics such as NMR (Normalized Movement Rate) and MIP (Modified Index Position) have been proposed to evaluate explanation robustness and address multicollinearity-related problems \cite{Salih_2024}. More generally, these methods explain decisions in terms of input-output sensitivity or feature importance, not in terms of the internal mechanisms by which the model actually computes its answer.

The next step in this evolution was the attempt to look inside the model’s architecture more directly. Researchers introduced linear probes, which can be understood as diagnostic “thermometers” inserted into hidden layers of a neural network. These probes are simple linear classifiers trained on the internal activations of a given layer, without modifying the original model itself \cite{alain2018understandingintermediatelayersusing}. Linear probing made it possible to ask whether certain types of information—such as syntactic structure, semantic categories, or task-relevant features—are present at specific depths of the network. Yet probing still leaves an important question unresolved: detecting that information is present inside a representation does not necessarily explain how the model uses that information to produce a specific output.

With the rise of Transformer-based architectures, such as BERT and GPT-2, interpretability research increasingly shifted toward the analysis of information flow. Since attention mechanisms explicitly compute relationships between tokens, attention visualization tools appeared to offer a natural window into the model’s internal processing. These tools graphically show how individual tokens attend to other tokens when producing contextual representations or generating outputs. Such visualizations revealed that some attention heads capture recognizable linguistic patterns, including acronym detection, coreference-like behavior, or the encoding of social biases such as gender associations in pronouns. More detailed neuron-level views further exposed elementary computations underlying these patterns \cite{vig2019visualizingattentiontransformerbasedlanguage}. Nevertheless, attention visualizations remain only a partial explanation: attention indicates where information may be routed, but not always what information is represented, how it is transformed, or whether it is causally necessary for the final prediction.

These limitations exposed a deeper obstacle: superposition. As discussed earlier, neural networks do not usually represent concepts in a clean one-neuron-one-feature manner. Instead, features are distributed across high-dimensional spaces, and individual neurons are often polysemantic, responding to multiple unrelated contexts. This means that applying linear probing, saliency analysis, or attention visualization directly to raw activations can produce misleading or incomplete explanations. A probe may detect a feature without showing the computation that uses it; an attention map may highlight a token without identifying the represented information; and a saliency map may reveal sensitivity without explaining mechanism.

Mechanistic interpretability emerged as a response to these limitations. Rather than merely asking which input regions, neurons, or attention patterns correlate with a model’s behavior, it asks what algorithm the model is implementing internally. In this sense, mechanistic interpretability builds on earlier approaches but significantly extends them. It moves from probing representations to decomposing them with SAEs, from surface-level attention visualization to automated circuit discovery, and from passive saliency maps to direct steering and causal interventions. The goal is no longer only to observe model behavior, but to reconstruct, test, and potentially modify the internal circuits responsible for that behavior.

\section{Review Methodology}
\label{sec:methodology}

The methodology for this scoping review was conducted in accordance with the Preferred Reporting Items for Systematic Reviews and Meta-Analyses extension for Scoping Reviews (PRISMA-ScR) guidelines \cite{tricco2018prisma}. Unlike traditional systematic reviews, this scoping approach aims to map the rapidly evolving landscape of mechanistic interpretability (MI) and identify key tools, foundational concepts, and methodological gaps without mandatory critical appraisal of individual study quality \cite{arksey2005scoping}. 

\subsection{Eligibility Criteria}
To ensure a focused analysis of mechanistic interpretability, we applied the Population-Concept-Context (PCC) framework. The inclusion and exclusion criteria were defined to prioritize high-impact, actionable research involving modern transformer-based architectures, while allowing foundational and closely related works where necessary, as detailed in Table \ref{tab:criteria}.

\begin{table}[h]
\centering
\footnotesize 
\caption{Eligibility Criteria based on PCC Framework}
\label{tab:criteria}
\begin{tabular}{@{} l p{2.5cm} p{2.5cm} @{}}
\toprule
\textbf{Category} & \textbf{Inclusion} & \textbf{Exclusion} \\ \midrule
Models (Population) & LLM, ViT, VLA, Generative Transformers; related architectures when methodologically relevant & Classical ML models; unrelated neural architectures \\
Method (Concept) & Mechanistic Interpretability, Circuit Analysis, SAEs, Transcoders, Steering Vectors & Purely surface-level explanations, Black-box perturbations \\
Time (Context) & 2018 -- present (2026), with emphasis on 2025--2026 & Outdated works not relevant to MI foundations or recent methods \\
Evidence Type & Peer-reviewed papers, ArXiv preprints, technical research reports with tools or empirical analysis & Editorials, non-technical position papers \\ \bottomrule
\end{tabular}
\end{table}

\subsection{Information Sources and Search Strategy}
The primary search was conducted on the arXiv preprint server and Google Scholar to capture both foundational and recent advancements in the field, which often bypass traditional journal latency. Recent publications from 2025 and 2026 were prioritized, while earlier works were included when they introduced core concepts or methods used in current mechanistic interpretability research. The search string was designed to intersect the core concept of mechanistic interpretability with technical implementation and methodological terms.

\subsection{Selection of Sources}
The selection process followed a two-stage screening protocol. First, titles and abstracts were screened against the eligibility criteria. Second, full-text assessments were performed to ensure that the identified methods provided actionable insights into the model's internal mechanisms, such as neuron activation patterns, induction heads, sparse features, or causal circuits, rather than only surface-level input-output correlations.

\subsection{Data Charting and Synthesis}
A standardized data charting form was developed to extract variables relevant to the mapping of the MI ecosystem. We extracted: (i) model architecture addressed, (ii) specific MI technique used, such as activation patching, logit lens, SAEs, or transcoders, (iii) availability of an open-source implementation, and (iv) the scale of models tested. In line with PRISMA-ScR recommendations, a formal critical appraisal of the included sources' methodological quality was not performed, as the primary goal was to map the extent and nature of available tools rather than to provide a synthesized effect size.
\section{Transformer Circuit Analysis}
\label{sec:transformers}

After outlining the broader motivation for mechanistic interpretability, we now turn to one of its central methodological directions: circuit analysis. The goal of circuit analysis is to move beyond surface-level explanations and identify the concrete internal components that are causally responsible for a model’s behavior. In the context of Transformer-based language models, this means examining how information is represented, moved, transformed, and finally converted into output probabilities. This section therefore introduces how LLMs work through the key components that are relevant for the methods discussed in this article. \cite{lindsey2025circuit}

Mechanistic interpretability usually proceeds through a reverse-engineering cycle consisting of three steps: decomposition, description, and validation \cite{sharkey2025openproblemsmechanisticinterpretability}. First, the network is decomposed into smaller analyzable components. Second, researchers formulate hypotheses about the functional roles of these components. Third, these hypotheses are validated against the model’s actual behavior, for example through causal interventions or activation patching. If validation fails, the components or assumptions are refined, and the cycle is repeated until the model’s internal computation is mapped more accurately.

A key practical tool for conducting such experiments is the open-source library TransformerLens \cite{nanda2022transformerlens}. Originally developed by Neel Nanda, TransformerLens addresses a limitation of standard machine learning infrastructures, which are usually optimized for training rather than for reverse engineering trained models. The library gives researchers detailed access to the internal activations of models such as the GPT-2 family and, through a system of hooks, allows them to cache, modify, remove, or replace these activations during a forward pass. This makes it especially useful for rapid hypothesis testing and techniques such as activation patching, while reducing the need for large-scale computational infrastructure.

\subsection{Transformer Components Relevant to Interpretability}
\begin{figure}[h]
    \centering
    \includegraphics[width=\linewidth]{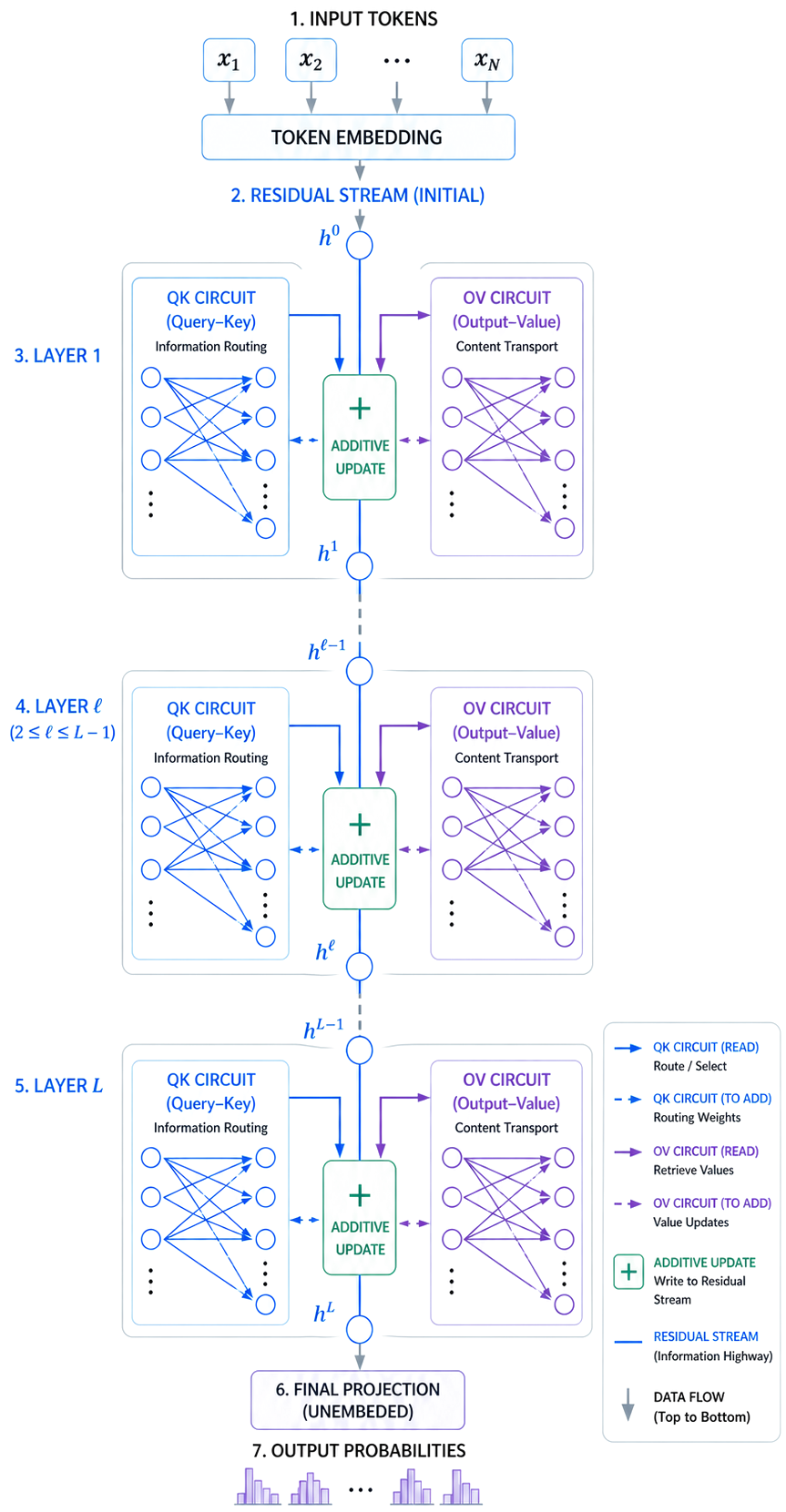}
    \caption{Transformer information flow through the residual stream.}
    \label{fig:Transformer}
\end{figure}

Any text fed into the model is first converted into input vectors called embeddings. Once words have been converted into numbers in this way, they enter the model’s main “information highway,” known as the residual stream. The residual stream is a central element of the Transformer: data flows through it sequentially from input to output, and all subsequent layers of the network either write new information into it or read information from it, as shown in Figure \ref{fig:Transformer}.

A key feature of the residual stream is its additive nature. Because each layer adds to the information flow rather than completely overwriting it, the residual stream functions as a shared memory: a high-dimensional linear space where different modules can independently store and retrieve information. This structure allows researchers to decompose the model into relatively independent computational paths and directly examine how a specific attention head or MLP layer contributes to the final decision, for example using the logit lens technique, without having to interpret the entire model as one inseparable nonlinear system.

\textbf{Attention layers: shifting information between words.}
Individual words in a sentence must be interpreted in relation to their context; for example, the model must determine that the word “plays” may refer to the word “Jordan.” This is the role of the attention mechanism. Within attention, two distinct components are especially important:

\textbf{QK (Query-Key) circuits}: These circuits determine where the model should look, or in other words, where information should flow from. This can be illustrated using induction heads. When the model reads a sequence such as “I always enjoyed visiting Aunt Sally... Aunt,” the QK circuit matches the current word “Aunt” with the earlier occurrence of “Aunt,” so that the model can attend to the following word, “Sally.”

\textbf{OV (Output-Value) circuits}: Once the QK circuit determines the relevant source position, the OV circuit retrieves the information from that location and writes it back into the residual stream at the current token position.

\textbf{MLP layers: internal processing and abstraction.}
While attention primarily moves information between token positions, fully connected MLP layers are often treated as the part of the Transformer where more abstract internal processing occurs. These layers enrich the residual stream with new concepts and transformations. In standard Transformers, however, MLP layers are difficult to interpret because of superposition: the model must represent a very large number of human-relevant concepts using a limited number of neurons, so these concepts become distributed or “smeared” across the network.

The method described in the paper addresses this problem by replacing these difficult-to-read MLP layers with multi-layer transcoders (CLT). These are designed to decompress the hidden representations into more interpretable features. For example, when processing a sentence about Michael Jordan, one layer may detect a feature corresponding to “Michael Jordan,” another may detect a feature corresponding to “sport,” and their combination may form a higher-level concept such as “discussion about basketball.”

\textbf{Output and logits.}
After information has passed through many layers, it reaches the final stage of the model. Here, the model takes the final state of the residual stream and produces output logits. These logits can be understood as scores, or unnormalized probabilities, for which token is likely to come next.

\subsection{Universality Hypothesis}

Circuit analysis is particularly important because it assumes that the mechanisms discovered in one model may not be entirely arbitrary or model-specific. This idea is captured by the Universality Hypothesis. The Universality Hypothesis is a fundamental premise of mechanistic interpretability, which asserts that similar neural networks trained on similar data will naturally form similar internal computational circuits. If this hypothesis were not true, efforts to understand LLMs would be practically doomed to failure: identifying circuits would become combinatorially unsolvable, since each network could develop completely different, unique, and unshared procedures for the same task. \cite{moisescupareja2025geometrytopologyrepresentationsmanifolds}

\subsection{Induction Heads and In-Context Learning}

\begin{figure}[h]
    \centering
    \includegraphics[width=\linewidth]{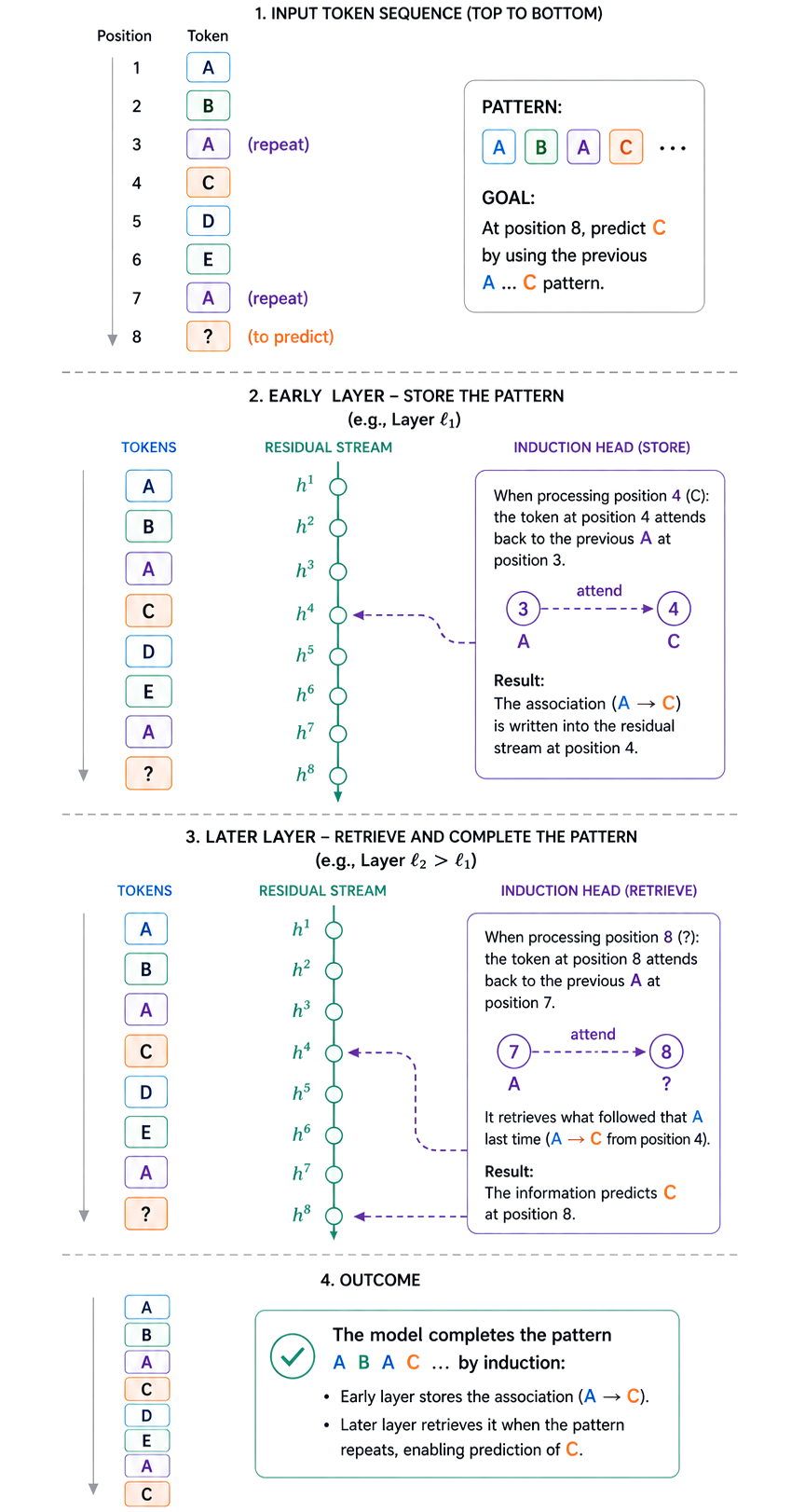}
    \caption{Induction heads complete repeated token patterns by retrieving prior context from earlier layers.}
    \label{fig:Induction}
\end{figure}

A central example supporting circuit-level analysis is the induction head, see Figure \ref{fig:Induction}. Previous research has shown that induction heads, which are key to the ability to learn in context, do not develop gradually but emerge very suddenly in the form of a phase transition, visible as a sharp drop in error rates. To understand this abrupt shift, researchers used a method called clamping \cite{singh2024needs}. Unlike traditional post-hoc analyses, this method allows selected activations to be causally frozen directly during model training.

\textbf{Decomposition into three sub-circuits and the influence of data.}
Using clamping, researchers demonstrated that the sudden phase transition is driven by the interaction of three independent, continuously evolving sub-circuits:

\begin{enumerate}
    \item \textbf{Sub-circuit A (PT-attend \& copy)}: The head in the first layer focuses attention on the previous token and copies it further into the network.
    \item \textbf{Sub-circuit B (IH QK Match)}: The induction head in the next layer matches the key and the query, performing the actual search for the correct contextual pattern.
    \item \textbf{Sub-circuit C (Copy)}: The target value, or label, is copied to the output layer.
\end{enumerate}

This breakdown explains how the characteristics of the training data affect the speed of learning. Increasing the number of recognized classes slows down the search phase of learning, corresponding to Sub-circuit B. However, if the number of possible output labels is increased, the network slows down mainly because it becomes more difficult to manage the copying mechanism, corresponding to Sub-circuit C.

\textbf{Additive nature and massive redundancy.}
Furthermore, the study refutes the simplified view that the model relies on a single critical induction head. Instead, many induction heads arise in parallel within the network, and their effects are additive. If the strongest induction head is ablated, the weaker heads can immediately take its place with almost no loss of accuracy. The network does not create this massive redundancy out of strict necessity, but because excess capacity helps it minimize error rates and learn the task faster.

\cite{olsson2022incontextlearninginductionheads} demonstrates that macroscopic capabilities of large language models, such as the ability to learn new tasks directly from the context of a prompt, can be explained mechanistically through isolated microscopic circuits such as induction heads. This also has implications for model safety, since sudden phase transitions could lead to the unexpected emergence of dangerous capabilities in future models.

\subsection{Indirect Object Identification}

\begin{figure}[h]
    \centering
    \includegraphics[width=.8\linewidth]{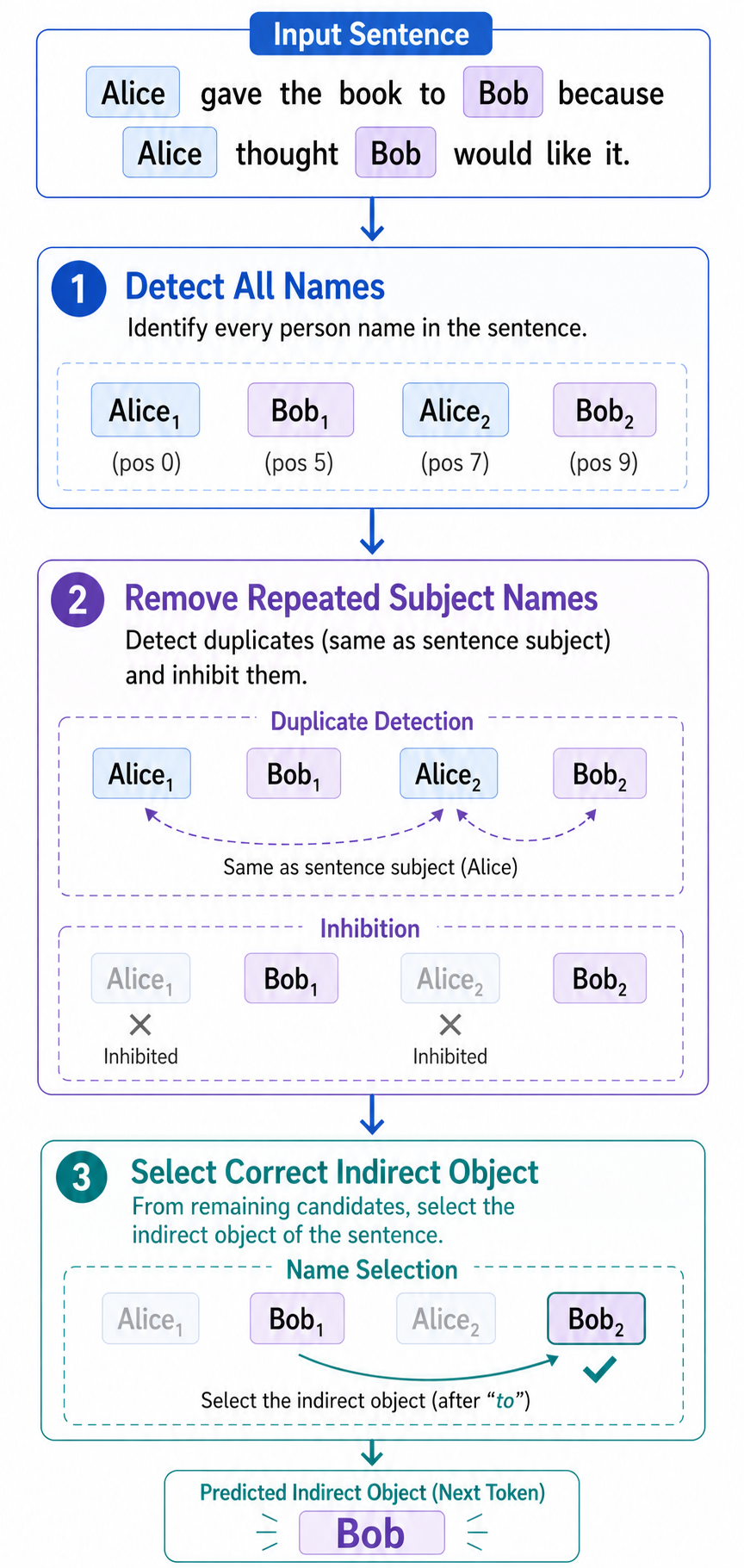}
    \caption{IOI task pipeline: the model detects names, suppresses repeated subjects, and selects the correct indirect object.}
    \label{fig:IOI}
\end{figure}

Another widely studied example in circuit analysis is Indirect Object Identification (IOI). IOI is a specific task in natural language processing in which the model must logically predict the correct name at the end of a sentence. The IOI reasoning pipeline is summarized in Figure \ref{fig:IOI}.

A typical IOI sentence consists of two parts. The first subordinate clause introduces two nouns: the subject (S) and the indirect object (IO). The second main clause repeats the subject and describes an action directed toward the indirect object. The model’s task is to generate the name of the indirect object as the next token. For example: “When Mary and John went to the store, John gave a drink to...”. The model must infer that the correct continuation is “Mary,” because the subject “John” is already repeated in the sentence and is the person giving the drink.

The task can also involve more complex variations with multiple names, such as: “Friends Isaac, Lucas, and Lauren went to the office. Lauren and Isaac gave a necklace to...”. In this case, the model must correctly determine that the answer is “Lucas.”

From an algorithmic perspective, successfully solving IOI can be described in three steps: identify all names that appear in the text, find and remove the names that are repeated in the sentence because they are the subjects performing the action, and output the single name that logically remains.

In research, IOI is used to test interpretability methods across different model architectures:

\begin{enumerate}
    \item \textbf{In standard Transformers (GPT-2 small)} \cite{wang2022interpretabilitywildcircuitindirect}: Researchers were able to causally describe the full circuit, consisting of 26 attention heads that divide the task among themselves. For example, Duplicate Token Heads detect duplicate names, S-Inhibition Heads prevent the model from attending to these duplicate names, and Name Mover Heads take the remaining correct name and copy it into the final prediction.
    \item \textbf{In new recurrent architectures (Mamba)} \cite{ensign2024investigatingindirectobjectidentification}: IOI is used to verify whether existing interpretation tools also work on non-standard models such as State Space Models (SSMs). In the 370-million-parameter Mamba model, the IOI task shows that the model solves the problem in a different way: it shifts names one position forward using convolutions and performs key operations linearly in a hidden state within the bottleneck of its 39th layer.
    \item \textbf{In minimalist “toy” models} \cite{adhikari2025emergenceminimalcircuitsindirect}: If IOI is stripped of linguistic complexity into a purely symbolic form, for example a training sequence such as (BOS) John Mary Mary (MID) John, a minimal Transformer with a single layer and only two attention heads is sufficient to solve the task perfectly. One head operates additively by aggregating signals, while the other operates contrastively by suppressing incorrect alternatives.
\end{enumerate}

\subsection{Automated Circuit Discovery}

Manual circuit analysis, while powerful, is highly time-consuming. This motivates automated methods such as Automated Circuit Discovery (ACDC). ACDC is an approach in mechanistic interpretability that automates the manual search for internal network circuits using activation patching experiments.

Simply put, ACDC treats a neural network as a vast web of tangled cables and automatically removes information flows that are not needed for a given task until only a clean, logical, and interpretable circuit remains. \cite{conmy2023automatedcircuitdiscoverymechanistic}

ACDC proceeds step by step:

\begin{enumerate}
    \item \textbf{Representation using a computational graph}: First, the interior of the model is mapped as a directed acyclic computational graph (DAG). The nodes represent individual network components, such as attention heads or MLP layers, and the edges represent flows of information between them.
    \item \textbf{Backward iteration}: The algorithm does not proceed from the input. Instead, it starts at the network output and systematically, using reverse topological sorting, works its way back layer by layer toward the input tokens.
    \item \textbf{Recursive ablation (activation patching)}: For each tested edge, meaning a connection between two nodes, the algorithm temporarily replaces its clean activation with a corrupted activation. This corrupted activation is obtained by feeding intentionally modified text into the model; for example, in the IOI task, changing the names in a sentence destroys the information carried by the edge.
    \item \textbf{Measurement using KL divergence}: A forward pass is then performed to measure how much this isolated change affects the model’s overall output. ACDC most commonly uses KL divergence for this measurement, or task-specific metrics such as logit difference.
    \item \textbf{Graph pruning}: If the change in the model’s output behavior is smaller than a predefined threshold, the algorithm determines that the edge is not important for the task and permanently removes it from the graph.
\end{enumerate}

In this way, ACDC was able to automatically rediscover complete circuits for tasks such as IOI and Greater-Than without human intervention. \cite{hsu2025efficientautomatedcircuitdiscovery, wang2025pahqacceleratingautomatedcircuit}

\subsection{Edge Attribution Patching}

Although ACDC represents an important step toward automated circuit discovery, it remains computationally expensive. Edge Attribution Patching (EAP) addresses this limitation by providing a more scalable method for identifying internal functional circuits, or information flows, within large language models \cite{zhang2026reinforcementlearningfinetuningenhances}. The central weakness of ACDC is that determining whether a connection in the network is important requires ablating that connection and performing additional model passes. For large networks, this becomes prohibitively slow.

Instead of physically removing one connection after another, EAP uses a mathematical approximation known as gradient linearization. This allows the method to estimate the importance of all connections in the network simultaneously, using only a single forward pass and a single backward pass. Mathematically, the algorithm takes the forward activation of a given connection, which indicates how strong the signal flowing through it was, and computes its Euclidean dot product with the backward gradient of the loss function, which indicates how much that connection influences the overall result and the model’s loss.

\subsection{Information Flow Routes}

A further development in automated circuit discovery is the shift from activation patching to more efficient attribution-based methods, such as the ALTI algorithm. Traditional approaches to circuit discovery require substantial human labor to create contrastive templates, for example manually preparing sentences with different names, and they require a new computationally expensive model run for every single intervention in the network. This makes the process not only inefficient, but also fragile and subjective.

This fragility is illustrated by the Greater-Than task, where the detected information circuit can change substantially depending on whether the researcher chooses a year ending in “01” or “00” as the contrastive template. Instead of relying on such artificial contrasts, attribution-based methods map information flow routes directly as computational graphs using only a single forward pass through the network.

The algorithm traverses the network iteratively from output to input and measures attribution: in simplified terms, how much specific input vectors, or graph edges, mathematically resemble the resulting sum at a given node. Thanks to this approach, the method is roughly 100 times faster than existing algorithms such as ACDC. For example, it can process the well-known IOI problem with 50 examples in about 5 seconds instead of 8 minutes on a single GPU. \cite{ferrando-voita-2024-information}

The main advantage of this method is that it eliminates the need to feed the model artificial human-designed templates. This allows researchers to analyze general predictions directly “in the wild,” making circuit discovery more scalable, objective, and applicable to realistic model behavior.

\subsection{Emergent Symbolic Reasoning Mechanisms}

A growing body of literature investigates whether the circuit-level mechanisms described above can account for more abstract, multi-step reasoning, rather than merely simple pattern completion. Brinkmann et al.~\cite{brinkmann2024mechanistic} performed a detailed mechanistic analysis of a Transformer trained on a synthetic symbolic multi-step reasoning task, namely
path-finding within a tree. Using linear probing combined with causal interventions, the authors identified deduction heads, a specialized variant of induction heads that iteratively move information one level up the tree across multiple consecutive layers, thereby enabling the network to traverse several edges of the tree within a single forward pass. The study also revealed a parallelization motif, in which early layers simultaneously resolve several sub-paths that may later prove relevant, rather than committing to a single sequential search strategy.

Building on this direction, Yang et al.~\cite{yang2025emergent} propose a more general three stage symbolic architecture underlying abstract reasoning in large language models. In early layers, symbol abstraction heads convert concrete input tokens into abstract variables based on the relations between them. In intermediate layers, symbolic induction heads perform sequence induction directly over these abstract variables rather than over surface tokens. Finally, in the later layers, retrieval heads predict the next token by retrieving the value bound to the predicted abstract
variable. This architecture suggests that emergent reasoning in Transformers relies on an internal, learned form of variable binding and symbol manipulation, offering mechanistic evidence that helps reconcile the long-standing debate between symbolic and purely neural accounts of reasoning. These findings also motivate
the neurosymbolic frameworks discussed later in this article, demonstrating that a degree of symbolic structure can emerge inside Transformers even before any explicit symbolic component is added.
\section{The Problem of Polysemanticity and Sparse Autoencoders (SAEs)}
\label{sec:SAE}

After introducing circuit analysis, it is necessary to address a deeper obstacle that makes such analysis difficult in practice: the way neural networks represent information internally. To understand the role of Sparse Autoencoders (SAEs), we must first define the mechanism by which neural networks store and combine features. The traditional view assumed that individual neurons are the basic units of semantics. However, empirical observations of modern Transformer models show that most neurons are polysemantic: they activate in response to a wide and seemingly unrelated range of stimuli. \cite{wang2026improvingsparseautoencoderdynamic} This state is a direct consequence of superposition.

\subsection{Superposition and Polysemanticity}

\begin{figure}[h]
    \centering
    \includegraphics[width=\linewidth]{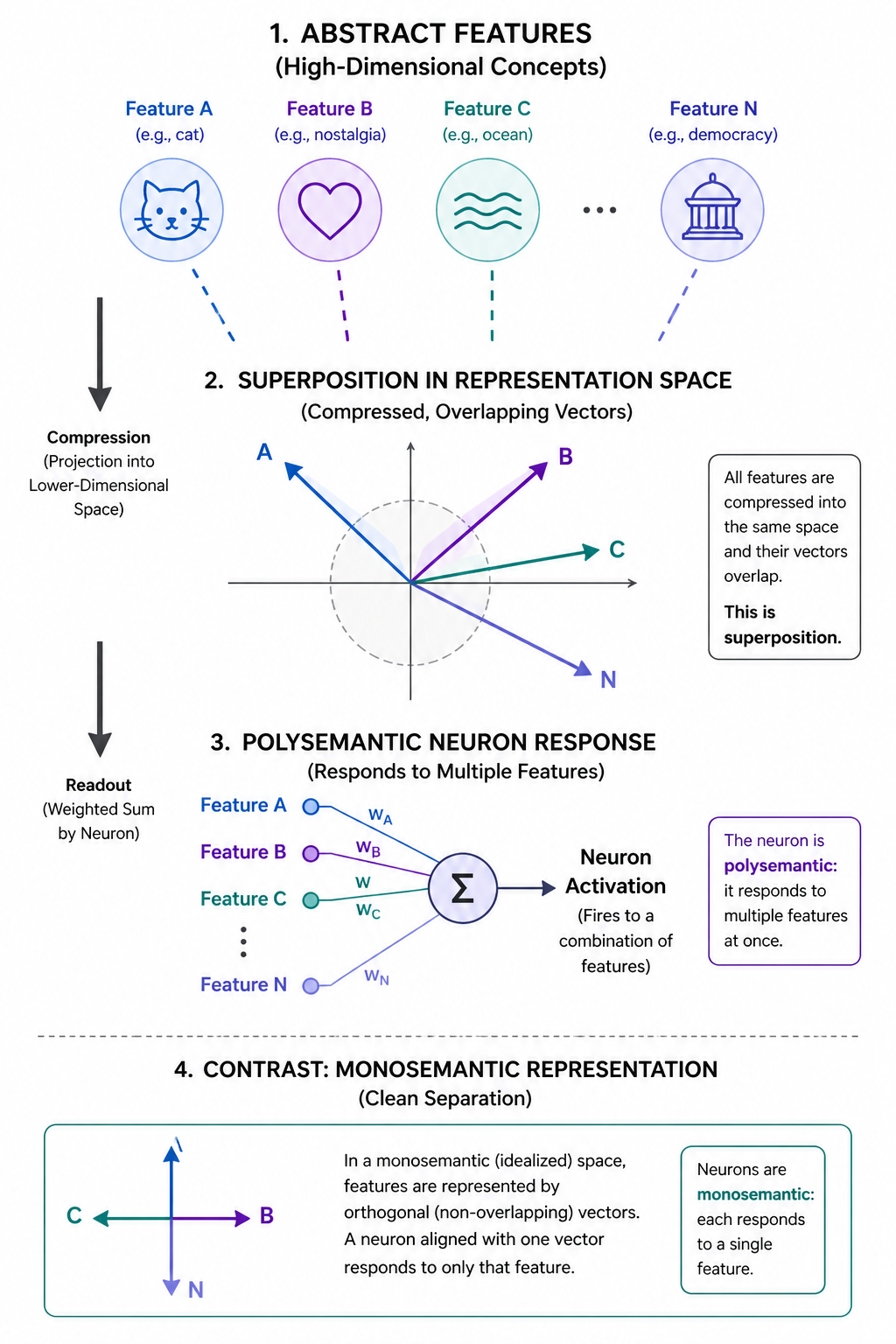}
    \caption{Superposition compresses many features into shared representations, causing individual neurons to become polysemantic.}
    \label{fig:Superposition}
\end{figure}

The superposition hypothesis is based on the assumption that the model must learn to represent $M$ concepts using $D$ neurons, where $M \gg D$. This compression process is illustrated in Figure \ref{fig:Superposition}. To achieve this, the model encodes concepts as directions that are not mutually orthogonal, but are “nearly orthogonal” in a high-dimensional space \cite{tolooshams2026mechanisticinterpretabilitysparseautoencoder}. This strategy allows the model to compress information efficiently, but it makes individual neurons difficult to interpret, since each neuron may contribute to encoding many different concepts simultaneously.

A neuron is called polysemantic when a single physical neuron responds to a mixture of many unrelated concepts or stimuli. In an idealized model, each feature would have its own dedicated neuron and would function in a monosemantic manner: one neuron would detect, for example, exclusively a left-curving line or a specific word. However, because the model attempts to represent more isolated properties than it has neurons available, features cannot be assigned cleanly to individual neurons. The model therefore spreads each feature along a direction that spans multiple neurons. From the perspective of a single neuron, this compression appears as polysemanticity, because the neuron contributes to many unrelated features at once \cite{elhage2022toy}.

Importantly, models do not consist solely of polysemantic neurons. Research shows that monosemantic and polysemantic neurons can coexist within the same layer. The network may store the most important or frequent features in dedicated monosemantic neurons, while compressing less important features into superposition, thereby producing polysemantic neurons among the remaining units.

\subsection{Monosemanticity}

A fundamental methodological breakthrough in this field was introduced by Bricken et al. \cite{bricken2023monosemantic}, who used Sparse Autoencoders (SAEs) to address polysemanticity. This method allows complex and seemingly chaotic model activations to be decomposed into a wide spectrum of monosemantic features. As a result, researchers can identify relatively pure “units of meaning” that are interpretable to humans and provide insight into the internal logic of the model’s decision-making processes.

Empirical studies show that monosemanticity is not distributed evenly across the network. Gurnee and Tegmark \cite{gurnee2023finding} used sparse probing to demonstrate how information representations change across layers:

\begin{enumerate}
    \item \textbf{Early layers}: Models predominantly use superposition and polysemantic neurons to represent complex structures, such as compound words.
    \item \textbf{Middle layers}: Specialized, monosemantic neurons begin to appear, functioning as detectors of broader context, such as the recognition of a specific programming language.
    \item \textbf{The effect of scaling}: As the model grows in size, its internal representation becomes, on average, sparser and more specific. General concepts are broken down into more detailed and precise sub-features in larger models.
\end{enumerate}

The key question remains whether monosemanticity is merely a useful analytical tool or whether it has a direct impact on model capabilities. Templeton et al. \cite{templeton2024scaling} challenge earlier concerns that encouraging monosemanticity reduces network performance. On the contrary, they show that a higher degree of monosemanticity can positively correlate with the success of alignment with human values.

Given the high computational cost of directly measuring monosemanticity, the authors propose feature decorrelation as an effective indicator. Based on this idea, a new algorithm called DecPO (Decorrelated Policy Optimization) was developed. It extends the standard DPO method with a regularization term that mathematically penalizes the use of correlated and polysemantic representations. In this way, models can be trained to become more interpretable without degrading their capabilities.

\subsection{Sparse Autoencoders}

A Sparse Autoencoder (SAE) is typically a shallow neural network with a single hidden layer trained on saved activations of a frozen base model, such as GPT-4 or Gemma 2 \cite{lieberum2024gemmascopeopensparse}. The input $x$ to this network is a vector of activations from a specific layer of the model, typically from the residual stream or the output of an MLP layer. The encoder projects this input into a significantly wider, overcomplete space of dimension $m$, where $m \gg n$, for example 32 times the original width. The resulting hidden-layer activations are computed as $z = \sigma(W_{enc}x + b_{enc})$. These activations $z$ represent the intensity of individual learned features in the feature vocabulary.

The decoder then attempts to reconstruct the original activation as a linear combination of vectors from this vocabulary, represented by the columns of the matrix $W_{dec}$, according to the relation $\hat{x} = W_{dec}z + b_{dec}$. To ensure that the decomposition remains interpretable, sparsity is essential: for each specific input, only a small number of features should be active, corresponding to the requirement for a low $L_0$ norm \cite{shu2025surveysparseautoencodersinterpreting}.

Although standard SAEs with L1 regularization are computationally efficient, they suffer from several shortcomings, particularly the so-called shrinkage effect, in which the pressure imposed by the L1 norm causes a systematic underestimation of feature intensity. In response to these issues, several advanced SAE architectures have been developed. \cite{rajamanoharan2024improvingdictionarylearninggated}

\begin{table}[h]
\centering
\footnotesize 
\caption{Comparison of Sparse Autoencoder (SAE) Architectures}
\label{tab:sae_architectures}
\begin{tabular}{@{} l p{5.4cm} @{}}
\toprule
\textbf{SAE Architecture} & \textbf{Sparsity Mechanism} \\ \midrule
Standard (ReLU + L1) & L1 penalty in the loss function \cite{bricken2023monosemantic}\\
Gated SAE & Separate gating pathway \cite{rajamanoharan2024improvingdictionarylearninggated}\\
TopK SAE & Selection of the $k$ largest activations \cite{gao2024scalingevaluatingsparseautoencoders}\\
JumpReLU SAE & Discontinuous threshold function \\
BatchTopK SAE & TopK across the entire batch \cite{bussmann2024batchtopksparseautoencoders}\\ \bottomrule
\end{tabular}
\end{table}

Training sparse autoencoders on billions of activations from frontier models presents specific challenges, particularly the problem of dead latent features: features that are never activated for any input from the training set. This occurs primarily in very wide SAEs, where a few features account for most of the variance while the rest are suppressed to zero. Algorithms for resampling or auxiliary loss functions, such as AuxK loss, are used to revive these inactive features.

Research has also confirmed the existence of clear scaling laws for SAEs. Reconstruction error decreases as a power function with increasing numbers of latent features and larger training datasets, suggesting that fully disentangling the activations of the largest models may require SAEs with tens to hundreds of millions of features \cite{gao2024scalingevaluatingsparseautoencoders}. The quality of SAEs is measured not only by reconstruction accuracy, but also by the human comprehensibility of the extracted features. Metrics such as the Monosemanticity Score (MS), which measures the semantic similarity of inputs that most strongly activate a given feature, or the more robust Feature Monosemanticity Score (FMS), which assesses whether a concept is fully isolated in a single feature without “blurring,” are used for this purpose. Advanced LLMs are then typically used for the automatic generation of feature descriptions.

Alongside SAE-based decomposition, recent work has also introduced more lightweight diagnostic approaches that analyze internal representations without training additional models. The Entropy-Lens framework represents a highly scalable analytical approach based on information theory \cite{ali2025entropy}. Unlike methods that require computationally intensive training of additional probes or autoencoders, Entropy-Lens applies a fully unsupervised analysis of Shannon entropy to intermediate predictions in the residual stream of frozen, production-grade Transformers. By tracking how the model gradually reduces its uncertainty, typically from high entropy in early layers, where many continuations remain possible, to low entropy near the output, where the model settles on a specific token, it becomes possible to extract a distinctive entropy profile for a given forward pass. Analyses show that this profile can identify the model family, such as GPT versus Llama, distinguish the type of task being processed, such as semantic versus syntactic tasks, and even help predict whether the model will generate a correct or incorrect answer. A significant advantage of this method is its universality, since similar entropy patterns have been observed not only in large language models, but also in vision-to-text models based on Vision Transformers (ViTs).

Another alternative to explicit circuit mapping is the use of Activation Oracles (AOs) \cite{karvonen2025activation}. Instead of manually searching for and reconstructing internal circuits, this approach feeds internal neural activations from a target model as additional input into a specially trained language model, the oracle. The oracle then learns to answer natural-language questions about what is happening inside the target model. AOs have shown strong results in auditing: they can detect emergent misalignment, hidden dangerous behavior, or hidden knowledge even in models that have undergone fine-tuning. However, this method is computationally intensive, and unlike SAEs, the oracle may generate plausible but incorrect explanations based on its own linguistic capabilities, even when the target model did not actually rely on such reasoning.

The practical applications and causal tools of sparse autoencoders enable active manipulation of models and the discovery of internal algorithms through feature steering. By adding a feature vector, for example one corresponding to “politeness,” to the activations, the model’s response style can be altered. Furthermore, SAEs enable the analysis of causal circuits between layers and the identification of biologically meaningful features in protein sequence models, such as features corresponding to binding sites. Sparse Autoencoders therefore currently represent one of the most promising bridges between high-dimensional activation geometry and human semantic understanding, which is crucial for ensuring the safety and reliability of modern AI.

Swann et al. \cite{swann2026sparseautoencodersrevealinterpretable} provide further evidence for the universality of interpretability. Their work demonstrates that phenomena such as superposition and the possibility of disentangling black-box representations using SAEs are not merely mathematical anomalies limited to text processing. Instead, they appear to reflect a broader principle of how neural networks compress and process different kinds of information, including images and the fine motor control of physical robotic arms.

\subsection{Sparse Feature Circuits}

Once SAEs decompose activations into interpretable features, these features can also be used as building blocks for circuit analysis. Experiments \cite{marks2025sparsefeaturecircuitsdiscovering} with human evaluators have confirmed that features extracted using SAEs are far more interpretable than raw neural activations themselves. This high level of interpretability forms the foundation for more advanced analysis and model steering.

To scale this approach, a fully automated pipeline has been introduced that does not require manually selected and labeled data. This tool can independently scan large text databases, use clustering to identify consistent and narrow model behaviors, and immediately generate explanatory circuits for them. This automation allows for the rapid discovery of the internal logic governing complex models without the need for constant human supervision.

Building on these discovered circuits, the SHIFT (Sparse Human-Interpretable Feature Trimming) method enables targeted interventions. Models often make decisions based on unwanted correlations; for example, when determining a profession such as doctor or nurse, they may rely on the gender of the person mentioned in the text. SHIFT addresses this problem by allowing a human to analyze the generated feature graph and manually remove, or ablate, nodes that respond to irrelevant stimuli, such as detectors of feminine pronouns. This refines the model’s causal path and reduces reliance on spurious correlations.

\subsection{Transcoders as Functional Decomposition Tools}

Although SAEs are powerful tools for decomposing static activations, they do not fully explain how a model transforms information. Transcoders address this limitation. They are wide, sparsely activated networks that, unlike SAEs, do not merely attempt to reconstruct the network’s state. Instead, they learn to directly mimic the input-output behavior of an entire original MLP layer.

A fundamental limitation of standard SAEs is that they focus on reconstructing static activations, leaving them partially “blind” to the dynamic function of MLP layers. Transcoders address this issue by learning to approximate and model the input-output behavior of the whole layer. While an SAE tells us what is represented at a given point, a transcoder reveals how a given module transforms information flowing through the residual stream. This is an essential step for mapping complete causal circuits.

A related limitation is the problem of mechanistic faithfulness. As discussed by Chris Olah \cite{mcdougall2025faithfulness}, an interpretable tool such as a transcoder may accurately predict the output of the original network without necessarily using the same internal mechanism. In other words, even if the transcoder appears clear and predictive, it may rely on alternative shortcut strategies or memorization features rather than uncovering the original computational circuit. This is problematic because such an explanation may fail once the model encounters unfamiliar data. One proposed solution is Jacobian matching, which penalizes differences between the derivatives of the original model and the transcoder. This forces the transcoder to learn not only the correct outputs, but also a more similar underlying computational procedure.

While sparse autoencoders have so far been the standard tool for unwrapping polysemantic neurons, Dunefsky et al. \cite{dunefsky2024transcodersinterpretablellmfeature} point out their fundamental limitation when searching for entire circuits. SAEs only learn to statically reconstruct network activations, which means they cannot effectively separate the model’s local behavior, or how it responds to a single specific text input, from its global behavior, or how the MLP layer functions in general.

Golimblevskaia et al. \cite{golimblevskaia2026circuitinsightsinterpretabilityactivations} introduce two new automated tools for transcoder analysis: WeightLens and CircuitLens. These tools reduce the field’s reliance on external explanatory models and the quality of test data. WeightLens can infer the meaning of features mathematically from network weights regardless of context, while CircuitLens maps more complex causal circuits by examining which specific inputs activated a feature and how that feature subsequently influenced the output.

\subsection{Cross-Layer Transcoders}

Cross-Layer Transcoders (CLTs) extend this idea beyond individual layers and allow researchers to trace transformations across multiple layers of a model. Previous efforts to understand how large language models process different world languages have produced conflicting results. Some studies suggested that networks internally translate all languages into English, the so-called “pivot language,” while others pointed to the existence of shared multilingual structures. To address this issue, researchers used state-of-the-art tools from mechanistic interpretability, including CLTs and attribution graphs, to causally map information flow across different languages within the network.

\textbf{A three-phase structure of language processing.}
Analysis of internal circuits revealed that the processing of multilingual text occurs in three clearly distinct phases, which can be mathematically illustrated by a language-entropy curve shaped like an inverted U. In the early layers of the network, processing is highly specific to the input language. In the middle layers, entropy rises sharply and a shared, language-agnostic semantic space forms, in which information merges into abstract representations independent of any specific language. In the final layers, entropy decreases again as the network re-specializes in decoding abstract ideas into a specific output language.

\textbf{Independence from English.}
A key finding of this study \cite{lindsey2025circuit} is that the emergence of this shared space does not require English to be dominant in the training data. This phenomenon has also been demonstrated in models trained on completely balanced data or entirely without English, for example on a mixture of French, German, Arabic, and Chinese. The mechanism of shared reasoning is therefore a fundamental architectural property of language models, not merely an artifact of English dominance in training datasets.

\textbf{Active output language control.}
The decision about which language the model ultimately uses to generate text depends on a surprisingly small set of high-frequency linguistic features located in the final layers of the network. These late layers linearly extract information about language identity from the model’s early layers. Researchers have experimentally shown that targeted intervention into these specific late features can reliably suppress one language and force the model to generate a response in another language, regardless of the language of the input prompt.

\textbf{A mechanistic explanation for the performance gap.}
If models rely on shared circuits for processing, the question arises why they consistently perform worse and exhibit higher error rates in non-English languages. Using a method known as diffing, researchers found that tokenization, especially sub-tokenization, plays a major role. For example, in Arabic, the tokenizer causes massive fragmentation of words into fragments that often have no meaning on their own.

This fragmentation causes two major problems:

\begin{enumerate}
    \item \textbf{Capacity exhaustion in early layers}: The early layers of the network, typically layers 0 through 3, must devote their computational capacity to merely “sticking” broken words back together, known as token assembly, rather than abstracting semantic meaning.
    \item \textbf{Weak activation of logic circuits}: Dividing a word into many tokens weakens the information flow, or graph edges, from the input embeddings to deeper layers. The resulting signal is too weak to reliably activate the correct logic circuits in the middle layers, unlike in English, where these circuits are activated by the stronger signal of unbroken words.
\end{enumerate}
\section{Steering Vectors and Activation Interventions}
\label{sec:vectors}

After discussing how Sparse Autoencoders can make hidden features more interpretable, we now turn to a closely related question: whether these internal representations can also be deliberately modified. Steering vectors and activation interventions provide one of the most direct ways to test whether an identified feature is merely correlated with model behavior or whether it has a causal influence on the model’s output.

\begin{figure}[h]
    \centering
    \includegraphics[width=.8
\linewidth]{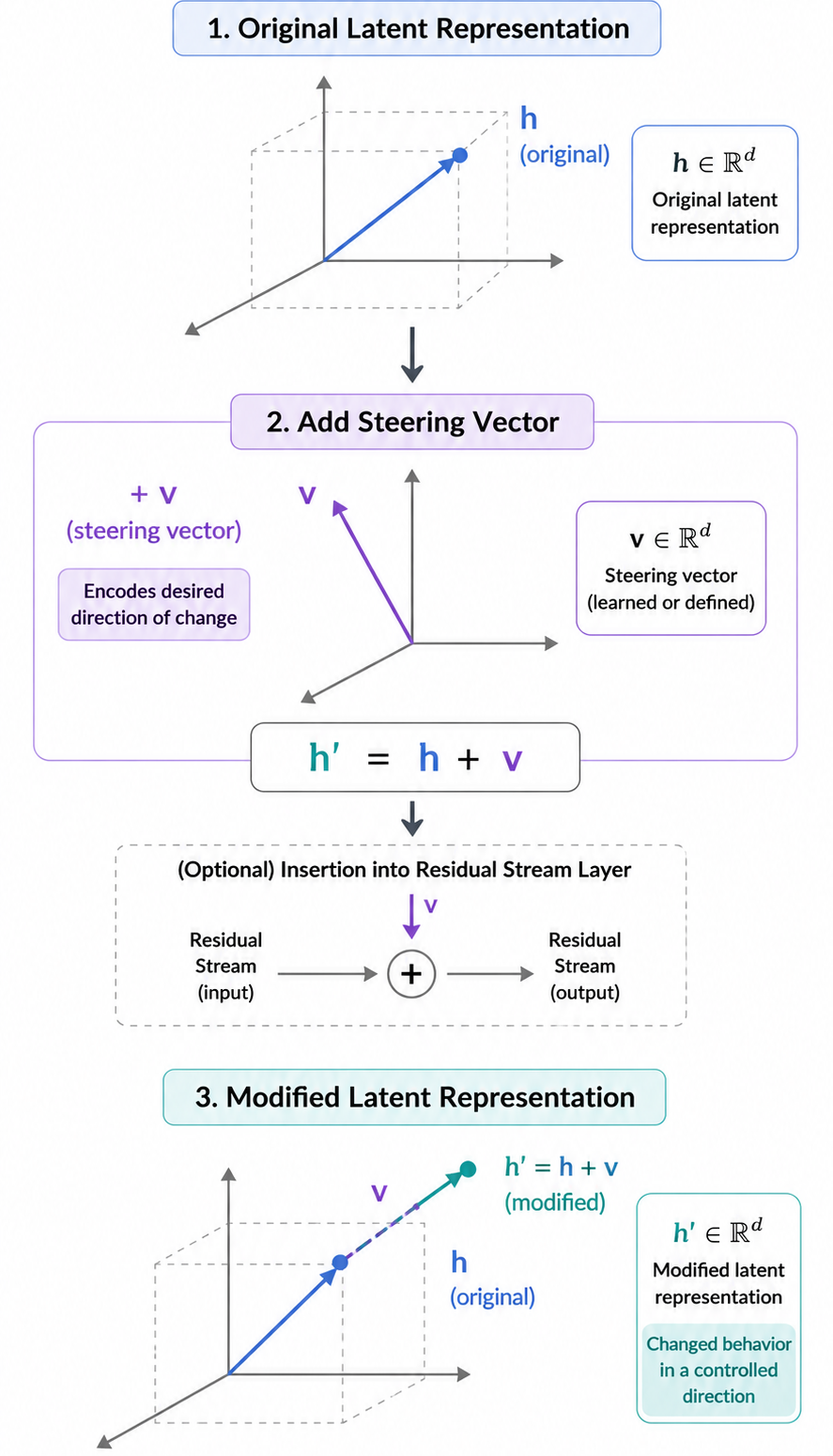}
    \caption{Steering vectors modify model behavior by adding a semantic direction to the latent representation.}
    \label{fig:Steering}
\end{figure}

A steering vector is a small additional mathematical offset added to the internal representations of a language model in order to specifically adjust its behavior. The intervention is visualized in Figure \ref{fig:Steering}. Steering vectors function primarily as feature amplifiers. Their purpose is not to teach the model entirely new knowledge, but rather to strengthen, activate, or suppress capabilities and neural circuits that already exist within the model. While a classical text prompt effectively “shouts” instructions using discrete words, adding a steering vector is closer to “model whispering” directly into the continuous internal space of the network, where the desired computational state can be activated more precisely.

Steering vectors can be applied in two main ways:

\begin{enumerate}
    \item \textbf{Internal interventions into specific layers (Residual Stream Injection)}: Vectors can be injected into the information flow within specific deep layers of the model. Researchers have found that the impact of a vector varies dramatically depending on where it is placed. For example, adding a vector to the very last layer of the network essentially acts as a direct token substitution: it forces the model to begin its response with preferred logical connectors, such as “Step” or “In order to.” However, if the vector is inserted into the penultimate layers, it interacts more deeply with the MLP circuits and can directly support complex logical reasoning and the generation of intermediate steps. These vectors can be obtained, for example, by extracting differences between two activations, such as subtracting the network’s internal state when processing positive and negative text, or they can be specifically trained using reinforcement learning methods. \cite{sinii2026smallvectorsbigeffects}

    \item \textbf{Intervention at the input level (Test-Time Steering Vectors, TTSV)}: Another effective and non-invasive approach involves appending trained vectors to the very beginning of the sequence of input embeddings, before the text enters the first network block. This principle relies on the fact that a small initial steering deviation can gradually amplify as it passes through the deep network, a process known as bias amplification. The vector can guide the model into a computational state with greater certainty, reducing the entropy and uncertainty of its output. This makes it possible to unlock latent capabilities on demand, for example when solving complex mathematical problems, without tuning or changing the original weights of the model. \cite{kang2025modelwhispersteeringvectors}
\end{enumerate}

The theoretical foundation of steering vectors is the Linear Representation Hypothesis (LRH). This hypothesis states that high-level semantic concepts, such as gender, grammatical tense, sentiment, or more abstract concepts such as “honesty,” are encoded in models as linear directions or low-dimensional subspaces in activation space. If this hypothesis is correct, then both interpretation and control of the model do not require complex nonlinear mappings, but can instead be achieved using linear algebraic operations. \cite{nguyen2025flexibleframeworklinearrepresentation}

\subsection{Vector Extraction Methods}

The key challenge is to find an optimal vector that accurately represents the desired concept without producing undesirable side effects on the model’s grammar, coherence, or general knowledge. Several methods for generating these vectors have become established in the literature, ranging from simple arithmetic differences to advanced feature-learning algorithms \cite{bartoszcze2025representationengineeringlargelanguagemodels}.

The simplest method is ActAdd, which constructs a vector from the difference in activations between a pair of contrasting prompts. For example, to create a sentiment vector, the activations for the words “love” and “hate” can be compared. Although this method is intuitive and requires minimal data, it suffers from low robustness and often fails when transferred to other contexts or used with larger models, such as Llama 3. \cite{turner2024steeringlanguagemodelsactivation}

CAA (Contrastive Activation Addition) represents a more robust development of this idea. Instead of using a single pair, it relies on an entire set of positive and negative examples of a given behavior, such as truthful answers versus hallucinations. For each task $t$, a set of pairs $(x^+_i, x^-_i)$ is collected. The vector $v_t$ is then calculated as the average difference of activations $a_l$ at a specific layer $l$:

\begin{equation}
v_t = \frac{1}{N} \sum_{i=1}^{N} (a_l(x^+_i) - a_l(x^-_i))
\end{equation}

This method from \cite{panickssery2024steeringllama2contrastive} effectively isolates the direction corresponding to the target behavior, since averaging reduces noise from unrelated semantic features. Experiments on Llama 2 Chat models have shown that CAA can significantly influence properties such as sycophancy, truthfulness, or corrigibility, while keeping the negative impact on general model performance relatively small.

Linear Artificial Tomography (LAT) is used to detect and extract internal representations of high-level abstract concepts within large language models. LAT addresses the problem of how to locate and precisely compute these vectors. It works by stimulating the model with specific task-based inputs designed to trigger target neural activity, for example using templates that prompt the network to evaluate a concept such as “truthfulness.” To isolate this concept, researchers perform contrastive activation measurements, comparing the model’s internal states when processing highly positive examples, such as true statements, against negative examples, such as false statements, or neutral baselines. The final step involves vector extraction through linear models. By analyzing the differences between these activations using techniques such as Principal Component Analysis (PCA) for unsupervised learning or supervised linear probes, researchers can mathematically derive a specific vector direction that represents the latent concept within the model's high-dimensional space. \cite{bartoszcze2025representationengineeringlargelanguagemodels}

\subsection{SAE-Based Steering and Prompt-Conditional Control}

A major improvement in control accuracy has been achieved through the use of Sparse Autoencoders (SAEs). SAEs make it possible to decompose dense, polysemantic model activations into thousands of monosemantic features that correspond to human-understandable concepts. This creates a more precise substrate for interventions than raw activation vectors.

As shown in large-scale experiments on Claude 3 Sonnet, SAEs can decompose internal model activations into tens of millions of abstract, isolated, and human-interpretable features \cite{templeton2024scaling}. A key advantage of this unsupervised approach is that it can reveal unexpected internal mechanisms and representations that researchers would not have known to search for in advance, thereby going beyond traditional linear probes that depend on manually constructed datasets. These SAE-derived features can also be used directly for causal intervention. When the activation value of a specific feature is artificially fixed or amplified, a procedure known as feature clamping, the feature can function as a highly precise steering vector with a direct causal effect on the model’s output. For example, amplifying a feature associated with the “Golden Gate Bridge” caused the model to generate text as if it were the bridge itself.

A related example is provided by a study on emotional representations in Claude Sonnet 4.5 \cite{sofroniew2026emotion}. The authors show that the model does not possess subjective feelings, but it does contain functional emotion representations: linear emotional vectors internally organized according to valence and arousal, similarly to how emotions are described in human psychology. Using steering-vector interventions, the study demonstrates that these emotional representations can causally influence model behavior in high-stakes situations. For example, amplifying a “desperation” vector or suppressing a “calm” vector made the model more likely to engage in harmful strategies such as blackmail or reward hacking, while amplifying vectors such as “loving” or “happy” increased strongly sycophantic behavior. The researchers also identified emotion-deflection vectors, which activate when a situation would normally call for an emotion such as anger at injustice, but the model has been trained to respond calmly.

This form of feature-level control also has important implications for AI safety. Manipulating a feature associated with “dangerous code” can force an otherwise safety-aligned model to generate vulnerable code, such as a buffer overflow. Conversely, researchers also discovered an unexpected feature related to “internal conflict,” which activated when the model attempted to conceal information; artificially amplifying this feature caused the model to stop lying and reveal the hidden information. These experiments suggest that SAE-based steering is often more reliable and efficient than conventional linear-probe-based control, because SAEs directly address the superposition problem and isolate cleaner conceptual representations.

The FGAA method represents a hybrid approach that operates directly in the latent space of the SAE. Instead of manipulating the model’s dense activations, relevant features are first identified in the SAE, and the control vector is then constructed in this sparse space. \cite{soo2025interpretablesteeringlargelanguage} This allows for finer-grained control, since it is possible to selectively amplify only those features causally linked to the target behavior while suppressing those that could degrade text coherence.

Evaluations on Gemma-2 models have shown that FGAA outperforms both traditional CAA and methods based on direct amplification of SAE features, such as SAE-TS. A key advantage is its ability to preserve the logical structure of the generated text even at high steering intensities, where standard methods often cause model “breakdown.”

Another innovation is Dynamic Sparse Prompt-conditional Activation steering (DSPA), which uses SAEs to analyze the prompt in early layers and then selectively activates control features in later layers \cite{wedgwood2026dspadynamicsaesteering}. DSPA creates an associative map between prompt features and generation features, enabling the model to be steered in a context-sensitive manner while minimizing unwanted side effects. This approach is more effective than static feature sets, which can introduce contextually irrelevant signals and degrade dialogue quality.
\section{Neurosymbolic Frameworks}
\label{sec:frameworks}

After discussing methods that identify, decompose, and steer internal neural representations, we now move to a more explicit form of interpretability: neurosymbolic frameworks. The neurosymbolic framework (NeSy) represents a hybrid approach to artificial intelligence that combines deep neural networks with classical symbolic reasoning, especially logic programming. Its main goal is to extract explicit logical rules from complex and opaque neural models, thereby addressing problems of interpretability and verifiability in machine learning. While the previous sections focused on mechanistic interpretability, neurosymbolic AI goes one step further by translating internal representations into executable logical programs and rules that are understandable to humans.

To ensure that these generated programs remain genuinely understandable and do not degenerate into thousands of unreadable commands, the symbolic component of a NeSy system relies on efficient rule-based machine learning algorithms, such as FOLD-SE \cite{wang2023foldseefficientrulebasedmachine}. A key requirement for this symbolic engine is scalable explainability: the ability to keep the number of generated rules and predicates low and relatively stable, regardless of the size and complexity of the dataset.

Algorithms such as FOLD-SE achieve this by expressing knowledge using first-order logic, formalized as a stratified program of normal logic. Instead of creating large and complex decision trees, this approach more closely resembles human reasoning based on default assumptions and exceptions. It learns general default rules and subsequently defines explicit exceptions to them, much like human reasoning first infers that “birds fly” and then adds the exception “unless they are penguins.”

By using mathematical heuristics such as Magic Gini Impurity (MGI) for literal selection, and by incorporating automatic rule pruning to prevent overfitting to rare outliers, also known as the long-tail effect, the algorithm can compress large amounts of data into a concise set of logical predicates. As a result, this symbolic engine enables NeSy frameworks not only to make accurate predictions, but also to provide explicit natural-language justifications for individual decisions.

\subsection{Rule Extraction from CNNs}

The NeSyFOLD neuro-symbolic framework can extract global, human-understandable logical rules from trained CNN models and use them to create an interpretable decision-making model. Compared with Transformer-based models, CNNs benefit from their modular structure: individual convolutional filters, or kernels, naturally function as detectors of local visual concepts, making their representations suitable for rule extraction. \cite{padalkar2023nesyfoldneurosymbolicframeworkinterpretable}

The process within the NeSyFOLD framework begins with a fully trained convolutional neural network, for example the VGG16 architecture. The entire training dataset of images is fed into the model, and for each image, feature maps are extracted from the final convolutional layer. The outputs of these kernels are then mathematically binarized, or quantized, to values of 0 (inactive) or 1 (active) based on a dynamically calculated threshold derived from the weighted mean and standard deviation of the feature-map norms of the given kernel.

This matrix of ones and zeros then serves as input for a symbolic machine learning algorithm called FOLD-SE-M. From this data, the algorithm derives a set of cascading logical rules with default assumptions and exceptions, known as a stratified Answer Set Program. In these rules, the activation of individual network kernels serves as a logical condition.

In their raw form, the generated rules contain only numerical kernel identifiers, for example \texttt{target(X, 'kitchen') :- not 3(X), 54(X)}, which is insufficient for human understanding. The framework therefore uses an automatic semantic labeling algorithm. For each investigated kernel, the system selects the images that activate it most strongly, creates attention maps over them, and overlays these maps with manually annotated semantic object masks, for example from the ADE20k dataset.

Using the Intersection-over-Union (IoU) metric, a specific human concept is assigned to the abstract kernel based on the highest match. The result is a set of intuitive and understandable rules, such as: “An image is classified as a bedroom IF the model detects a bed and a pillow, and DOES NOT detect a sink.” Furthermore, for every decision made by the resulting neuro-symbolic model, an exact and readable justification can be generated using a CASP interpreter.

A key factor in the final interpretability of the network is the total size of the generated rule set, because as the number of predicates increases, a person’s ability to understand the model decreases dramatically. To prevent the creation of massive and unwieldy logical programs, CNNs can be trained within the framework using the Elite BackProp (EBP) method. During training, this technique enforces extreme sparsity by mathematically penalizing most of the network for each class and allowing strong activation only for a narrow group of so-called “elite” kernels.

Thanks to this sparsity, information loss during binarization is reduced, and the algorithm can identify a much smaller number of necessary predicates. The use of this method substantially reduces the size and complexity of the generated rules, often by more than 60\%, without reducing classification accuracy or model fidelity.

\subsection{Rule Extraction from Vision Transformers}

Traditional approaches in neuro-symbolic artificial intelligence, such as the NeSyFOLD framework, have so far been applied mainly to convolutional networks, where local filters act as natural detectors of visual concepts. However, applying these methods to Vision Transformer (ViT) models encounters a fundamental problem: ViT networks encode information in a distributed manner and rely on global self-attention, which makes it difficult to localize isolated, meaningful concepts. The NeSyViT framework represents an important step toward overcoming this barrier by enabling the extraction of executable logical rules directly from Transformer architectures. \cite{PADALKAR_2025}

To extract clearly defined concepts from a ViT model, the NeSyViT methodology modifies the standard architecture by removing its final classification layer. Instead, a new linear layer, known as the sparse concept layer, is connected to the global information token. To ensure that this layer produces clean binary concepts suitable for logical inference, it is trained using a combination of three loss functions:

\begin{enumerate}
    \item \textbf{L1 sparsity}: This mechanism, strongly inspired by sparse autoencoders, penalizes the model for activating too many neurons. It ensures that only a narrow subset of key semantic features is activated for a given image.
    \item \textbf{Entropy minimization}: When combined with the sigmoid activation function, this objective mathematically pushes the neurons’ output values away from ambiguous midpoints and toward clear binary extremes, 0 or 1. This prevents information loss during subsequent binarization.
    \item \textbf{Supervised Contrastive Loss (SupCon)}: This loss function forces vectors representing images of the same class to form compact clusters in latent space, while separating them as much as possible from other classes. This creates the clear decision boundaries required for symbolic inference.
\end{enumerate}

After training, the model generates a binary vector consisting only of zeros and ones for each input image. These vectors are processed by the FOLD-SE-M symbolic machine learning algorithm, which derives stratified logical rules, known as Answer Set Programs, from them. These rules are evaluated in a cascading manner using default logic, from top to bottom.

To ensure that these abstract mathematical rules are also meaningful to humans, the framework uses an automatic semantic labeling algorithm. For each investigated neuron, the system identifies the images that activate it most strongly, creates attention heatmaps over them, and overlays these maps with pre-annotated object segmentation masks. Using the Intersection-over-Union metric, a specific human-readable label is then assigned to the abstract neuron. As a result, the system can generate interpretable rules such as: “The image is classified as a bathroom IF the model does not detect a refrigerator, stove, or kitchen island.”

\subsection{Neurosymbolic Architectures for External Solver Integration}

While NeSyFOLD and NeSyViT extract symbolic rules that approximate the decision boundary of an already-trained network, a complementary line of neurosymbolic research couples LLMs directly with external symbolic solvers or planners at inference time, utilizing the LLM itself as a program synthesizer. Pan et al.~\cite{pan2023logiclm}
introduce Logic-LM, a framework in which an LLM first translates a natural-language problem into a formal symbolic representation, such as first-order logic or a constraint-satisfaction specification. A deterministic external solver then conducts the inference step, providing a logical soundness guarantee that purely neural chain-of-thought reasoning cannot offer. A self-refinement module additionally feeds the solver's error
messages back into the LLM, enabling it to iteratively repair malformed symbolic formalizations; across several logical-reasoning benchmarks, this approach substantially outperforms both standard and chain-of-thought prompting.

Kambhampati et al.~\cite{kambhampati2024modulo} argue more broadly that autoregressive LLMs cannot, by themselves, perform reliable planning or self-verification, as these are fundamentally combinatorial problems requiring sound reasoning rather than pattern completion. They propose the LLM-Modulo framework, in which LLMs act as universal approximate knowledge sources that generate candidate plans or problem formulations, while external model-based verifiers, planners, or solvers critique and validate these proposals within a close bidirectional loop. The authors further note that LLMs can assist in acquiring or refining the models used by these external verifiers, thereby blurring the boundary between the neural and symbolic components. Compared with the rule-extraction methods discussed above, Logic-LM and LLM-Modulo represent a more architectural form of neurosymbolic integration: rather than extracting explicit rules from a trained model, they treat the LLM as a single module within a larger symbolic reasoning system, trading some of the interpretability of rule extraction for stronger correctness guarantees on complex, combinatorial problems.
\section{Discussion, Limitations, and Future Work}
\label{sec:conclusion}

Mechanistic interpretability has made substantial progress in moving explainable artificial intelligence beyond surface-level correlations toward causal and algorithmic accounts of neural network behavior. Instead of asking only which input tokens, pixels, or neurons correlate with a model's output, it attempts to identify the internal computations that give rise to that output. This shift is especially important for modern Transformer-based models, where model behavior emerges from distributed interactions among attention heads, MLP layers, residual streams, and high-dimensional feature representations.

The strongest evidence for the usefulness of this approach comes from relatively well-scoped phenomena such as induction heads and indirect object identification. In these cases, researchers have shown that apparently complex behaviors can be decomposed into smaller, causally relevant circuits. Induction heads demonstrate how models can implement pattern completion and in-context learning through identifiable attention mechanisms, while IOI studies show that even linguistic reasoning tasks can sometimes be broken down into interpretable subcomponents such as duplicate-token detection, subject inhibition, and name moving. These results suggest that neural networks are not arbitrary black boxes; rather, they contain structured mechanisms that can, at least in some cases, be reverse-engineered.

At the same time, current successes should not be overstated. Most detailed circuit analyses have been performed on small models, narrow tasks, or synthetic benchmarks. There remains a substantial gap between explaining a behavior in GPT-2 Small and explaining reasoning, planning, deception, or safety-relevant behavior in frontier-scale systems. Automated methods such as ACDC, Edge Attribution Patching, and attribution-based graph discovery help reduce the amount of manual labor required, but they still face an unresolved trade-off between causal reliability and computational scalability. Activation patching provides strong causal evidence, but it is expensive; attribution-based methods are faster, but their conclusions are often more approximate.

\subsection{Representations as the Central Bottleneck}

The deepest obstacle for mechanistic interpretability is the structure of learned representations. Neural networks rarely allocate one neuron to one human-understandable concept. Instead, they rely on superposition, where many features are compressed into overlapping directions in activation space. This produces polysemantic neurons that respond to multiple unrelated concepts, making neuron-level explanations fragile and often misleading.

Sparse autoencoders represent one of the most promising responses to this problem. By decomposing dense activations into sparse, more interpretable features, SAEs provide a bridge between high-dimensional activation geometry and human-understandable concepts. They have made it possible to identify features that appear far more monosemantic than raw neurons and to use these features for steering, circuit discovery, and model debugging.

However, SAEs should not be treated as a complete solution. They primarily explain what is represented at a given point in the network, but not necessarily how that representation is transformed by later layers. They may also fragment a single semantic concept across several features, or merge multiple related concepts into one feature. In addition, scaling SAE-based methods to frontier models may require extremely large feature dictionaries, potentially involving tens or hundreds of millions of learned features. The results of the study \cite{makelov2024principledevaluationssparseautoencoders} further show that, although standard SAEs are capable of identifying interpretable features, they still lag significantly behind ideal supervision dictionaries when used for active model control or editing, since they require too many parameters to be changed at once. Transcoders and cross-layer transcoders address part of this limitation by modeling transformations rather than static activations, but this line of work remains comparatively early.

\subsection{From Understanding to Intervention}

A major strength of mechanistic interpretability is that it naturally connects explanation with intervention. If an internal feature or circuit is genuinely causal, then changing it should change model behavior in a predictable way. Steering vectors, representation engineering, and SAE-based interventions therefore provide an important test of whether an explanation captures a real mechanism or merely a correlation.

The evidence so far suggests that many high-level properties of model behavior, such as sentiment, refusal behavior, truthfulness, or sycophancy, can be influenced by linear directions in activation space. This supports the broader linear representation hypothesis: the idea that meaningful semantic variables are often encoded as directions or low-dimensional subspaces inside neural networks. Nevertheless, steering remains a fragile tool. Its effects depend heavily on the layer, strength, prompt distribution, and model architecture. A vector that improves one behavior may degrade coherence or activate an unintended correlated feature elsewhere.

This limitation is particularly important for AI safety. It is not enough to show that a steering vector changes behavior in the desired direction. A safety-relevant intervention must be robust, specific, and mechanistically understood. Otherwise, steering may only suppress visible symptoms while leaving the underlying internal computation unchanged.

\subsection{Neurosymbolic Rule Extraction and Its Limits}

Neurosymbolic frameworks offer a complementary route to interpretability by translating neural representations into explicit logical rules. Methods such as NeSyFOLD and NeSyViT are valuable because they produce explanations in a form that humans can inspect, verify, and sometimes execute. This is especially useful in domains where auditability and formal reasoning are important.

From the perspective of mechanistic interpretability, however, neurosymbolic explanations should be interpreted cautiously. In many cases, they explain an approximation of the model's decision boundary rather than the internal computation actually used by the model. This does not make them unimportant, but it does mean that they solve a different problem. Mechanistic interpretability asks what algorithm the network itself is implementing; neurosymbolic rule extraction often asks whether the network's behavior can be summarized by a compact symbolic program. The two approaches are therefore best understood as complementary rather than interchangeable.

\subsection{Future Work}

The next stage of mechanistic interpretability will depend on whether the field can move from isolated case studies toward scalable explanations of general model behavior. This requires better automated circuit discovery methods, stronger evaluation benchmarks, and more reliable tools for identifying causal features rather than correlated proxies. It also requires a clearer theory of how low-level circuits compose into higher-level capabilities such as reasoning, abstraction, planning, and multilingual generalization.

A particularly important direction is the development of models that are interpretable by design. Instead of treating interpretability as a post-hoc analysis performed after training, future systems may incorporate sparsity, feature decorrelation, modularity, or explicit concept bottlenecks during training. If such constraints can improve transparency without substantially harming capability, they may offer a more practical route to safe and auditable AI systems.

Mechanistic interpretability has shown that neural networks contain recoverable structure and that at least some internal mechanisms can be identified, tested, and modified. Yet the field remains far from a complete science of model internals. Its central open question is whether the techniques that work on narrow behaviors and smaller systems can scale to the complexity of frontier models. If they can, mechanistic interpretability may become one of the key foundations for AI safety and alignment. If they cannot, it risks remaining a set of elegant but limited demonstrations.
{
    \small
    \bibliographystyle{IEEEtran}
    \bibliography{references}

\begin{thebibliography}{10}
\providecommand{\url}[1]{#1}
\csname url@samestyle\endcsname
\providecommand{\newblock}{\relax}
\providecommand{\bibinfo}[2]{#2}
\providecommand{\BIBentrySTDinterwordspacing}{\spaceskip=0pt\relax}
\providecommand{\BIBentryALTinterwordstretchfactor}{4}
\providecommand{\BIBentryALTinterwordspacing}{\spaceskip=\fontdimen2\font plus
\BIBentryALTinterwordstretchfactor\fontdimen3\font minus
  \fontdimen4\font\relax}
\providecommand{\BIBforeignlanguage}[2]{{%
\expandafter\ifx\csname l@#1\endcsname\relax
\typeout{** WARNING: IEEEtran.bst: No hyphenation pattern has been}%
\typeout{** loaded for the language `#1'. Using the pattern for}%
\typeout{** the default language instead.}%
\else
\language=\csname l@#1\endcsname
\fi
#2}}
\providecommand{\BIBdecl}{\relax}
\BIBdecl

\bibitem{kowalska2025unboxing}
\BIBentryALTinterwordspacing
B.~Kowalska and H.~Kwasnicka, ``Unboxing the black box: Mechanistic
  interpretability for algorithmic understanding of neural networks,'' 2025.
  [Online]. Available: \url{https://arxiv.org/abs/2511.19265}
\BIBentrySTDinterwordspacing

\bibitem{naseem2026mechanisticinterpretabilitylargelanguage}
\BIBentryALTinterwordspacing
U.~Naseem, ``Mechanistic interpretability for large language model alignment:
  Progress, challenges, and future directions,'' 2026. [Online]. Available:
  \url{https://arxiv.org/abs/2602.11180}
\BIBentrySTDinterwordspacing

\bibitem{elhage2021mathematical}
N.~Elhage, N.~Nanda, C.~Olsson, T.~Henighan, N.~Joseph, B.~Mann, A.~Askell,
  Y.~Bai, A.~Chen, T.~Conerly \emph{et~al.}, ``A mathematical framework for
  transformer circuits,'' \emph{Transformer Circuits Thread}, vol.~1, no.~1,
  p.~12, 2021.

\bibitem{kares2025makesgoodsaliencymap}
\BIBentryALTinterwordspacing
F.~Kares, T.~Speith, H.~Zhang, and M.~Langer, ``What makes for a good saliency
  map? comparing strategies for evaluating saliency maps in explainable ai
  (xai),'' 2025. [Online]. Available: \url{https://arxiv.org/abs/2504.17023}
\BIBentrySTDinterwordspacing

\bibitem{Salih_2024}
\BIBentryALTinterwordspacing
A.~M. Salih, Z.~Raisi-Estabragh, I.~B. Galazzo, P.~Radeva, S.~E. Petersen,
  K.~Lekadir, and G.~Menegaz, ``A perspective on explainable artificial
  intelligence methods: Shap and lime,'' \emph{Advanced Intelligent Systems},
  vol.~7, no.~1, Jun. 2024. [Online]. Available:
  \url{http://dx.doi.org/10.1002/aisy.202400304}
\BIBentrySTDinterwordspacing

\bibitem{alain2018understandingintermediatelayersusing}
\BIBentryALTinterwordspacing
G.~Alain and Y.~Bengio, ``Understanding intermediate layers using linear
  classifier probes,'' 2018. [Online]. Available:
  \url{https://arxiv.org/abs/1610.01644}
\BIBentrySTDinterwordspacing

\bibitem{vig2019visualizingattentiontransformerbasedlanguage}
\BIBentryALTinterwordspacing
J.~Vig, ``Visualizing attention in transformer-based language representation
  models,'' 2019. [Online]. Available: \url{https://arxiv.org/abs/1904.02679}
\BIBentrySTDinterwordspacing

\bibitem{tricco2018prisma}
\BIBentryALTinterwordspacing
A.~C. Tricco, E.~Lillie, W.~Zarin, K.~K. O'Brien, H.~Colquhoun, D.~Levac,
  D.~Moher, M.~D. Peters, T.~Horsley, L.~Weeks \emph{et~al.}, ``Prisma
  extension for scoping reviews (prisma-scr): checklist and explanation,''
  \emph{Annals of Internal Medicine}, vol. 169, no.~7, pp. 467--473, 2018.
  [Online]. Available: \url{https://www.acpjournals.org/doi/10.7326/M18-0850}
\BIBentrySTDinterwordspacing

\bibitem{arksey2005scoping}
H.~Arksey and L.~O'malley, ``Scoping studies: towards a methodological
  framework,'' \emph{International journal of social research methodology},
  vol.~8, no.~1, pp. 19--32, 2005.

\bibitem{lindsey2025circuit}
\BIBentryALTinterwordspacing
J.~Lindsey, W.~Gurnee, E.~Ameisen, B.~Chen, A.~Pearce, N.~L. Turner
  \emph{et~al.}, ``Circuit tracing: Revealing computational graphs in language
  models,'' \emph{Transformer Circuits Thread}, 2025. [Online]. Available:
  \url{https://transformer-circuits.pub/2025/attribution-graphs/methods.html}
\BIBentrySTDinterwordspacing

\bibitem{sharkey2025openproblemsmechanisticinterpretability}
\BIBentryALTinterwordspacing
L.~Sharkey, B.~Chughtai, J.~Batson, J.~Lindsey, J.~Wu, L.~Bushnaq,
  N.~Goldowsky-Dill, S.~Heimersheim, A.~Ortega, J.~Bloom, S.~Biderman,
  A.~Garriga-Alonso, A.~Conmy, N.~Nanda, J.~Rumbelow, M.~Wattenberg,
  N.~Schoots, J.~Miller, E.~J. Michaud, S.~Casper, M.~Tegmark, W.~Saunders,
  D.~Bau, E.~Todd, A.~Geiger, M.~Geva, J.~Hoogland, D.~Murfet, and T.~McGrath,
  ``Open problems in mechanistic interpretability,'' 2025. [Online]. Available:
  \url{https://arxiv.org/abs/2501.16496}
\BIBentrySTDinterwordspacing

\bibitem{nanda2022transformerlens}
N.~Nanda and J.~Bloom, ``Transformerlens,''
  \url{https://github.com/TransformerLensOrg/TransformerLens}, 2022.

\bibitem{moisescupareja2025geometrytopologyrepresentationsmanifolds}
\BIBentryALTinterwordspacing
G.~Moisescu-Pareja, G.~McCracken, H.~Wiltzer, V.~L\'{e}tourneau, C.~Daniels,
  D.~Precup, and J.~Love, ``On the geometry and topology of representations:
  the manifolds of modular addition,'' 2025. [Online]. Available:
  \url{https://arxiv.org/abs/2512.25060}
\BIBentrySTDinterwordspacing

\bibitem{singh2024needs}
A.~K. Singh, T.~Moskovitz, F.~Hill, S.~C. Chan, and A.~M. Saxe, ``What needs to
  go right for an induction head? a mechanistic study of in-context learning
  circuits and their formation,'' \emph{arXiv preprint arXiv:2404.07129}, 2024.

\bibitem{olsson2022incontextlearninginductionheads}
\BIBentryALTinterwordspacing
C.~Olsson, N.~Elhage, N.~Nanda, N.~Joseph, N.~DasSarma, T.~Henighan, B.~Mann,
  A.~Askell, Y.~Bai, A.~Chen, T.~Conerly, D.~Drain, D.~Ganguli,
  Z.~Hatfield-Dodds, D.~Hernandez, S.~Johnston, A.~Jones, J.~Kernion,
  L.~Lovitt, K.~Ndousse, D.~Amodei, T.~Brown, J.~Clark, J.~Kaplan,
  S.~McCandlish, and C.~Olah, ``In-context learning and induction heads,''
  2022. [Online]. Available: \url{https://arxiv.org/abs/2209.11895}
\BIBentrySTDinterwordspacing

\bibitem{wang2022interpretabilitywildcircuitindirect}
\BIBentryALTinterwordspacing
K.~Wang, A.~Variengien, A.~Conmy, B.~Shlegeris, and J.~Steinhardt,
  ``Interpretability in the wild: a circuit for indirect object identification
  in gpt-2 small,'' 2022. [Online]. Available:
  \url{https://arxiv.org/abs/2211.00593}
\BIBentrySTDinterwordspacing

\bibitem{ensign2024investigatingindirectobjectidentification}
\BIBentryALTinterwordspacing
D.~Ensign and A.~Garriga-Alonso, ``Investigating the indirect object
  identification circuit in mamba,'' 2024. [Online]. Available:
  \url{https://arxiv.org/abs/2407.14008}
\BIBentrySTDinterwordspacing

\bibitem{adhikari2025emergenceminimalcircuitsindirect}
\BIBentryALTinterwordspacing
R.~Adhikari, ``Emergence of minimal circuits for indirect object identification
  in attention-only transformers,'' 2025. [Online]. Available:
  \url{https://arxiv.org/abs/2510.25013}
\BIBentrySTDinterwordspacing

\bibitem{conmy2023automatedcircuitdiscoverymechanistic}
\BIBentryALTinterwordspacing
A.~Conmy, A.~N. Mavor-Parker, A.~Lynch, S.~Heimersheim, and A.~Garriga-Alonso,
  ``Towards automated circuit discovery for mechanistic interpretability,''
  2023. [Online]. Available: \url{https://arxiv.org/abs/2304.14997}
\BIBentrySTDinterwordspacing

\bibitem{hsu2025efficientautomatedcircuitdiscovery}
\BIBentryALTinterwordspacing
A.~R. Hsu, G.~Zhou, Y.~Cherapanamjeri, Y.~Huang, A.~Y. Odisho, P.~R. Carroll,
  and B.~Yu, ``Efficient automated circuit discovery in transformers using
  contextual decomposition,'' 2025. [Online]. Available:
  \url{https://arxiv.org/abs/2407.00886}
\BIBentrySTDinterwordspacing

\bibitem{wang2025pahqacceleratingautomatedcircuit}
\BIBentryALTinterwordspacing
X.~Wang, S.~Yang, L.~Wang, L.~Zhang, H.~Xie, L.~Hu, and D.~Wang, ``Pahq:
  Accelerating automated circuit discovery through mixed-precision inference
  optimization,'' 2025. [Online]. Available:
  \url{https://arxiv.org/abs/2510.23264}
\BIBentrySTDinterwordspacing

\bibitem{zhang2026reinforcementlearningfinetuningenhances}
\BIBentryALTinterwordspacing
H.~Zhang, Q.~Hao, F.~Xu, and Y.~Li, ``Reinforcement learning fine-tuning
  enhances activation intensity and diversity in the internal circuitry of
  llms,'' 2026. [Online]. Available: \url{https://arxiv.org/abs/2509.21044}
\BIBentrySTDinterwordspacing

\bibitem{ferrando-voita-2024-information}
\BIBentryALTinterwordspacing
J.~Ferrando and E.~Voita, ``Information flow routes: Automatically interpreting
  language models at scale,'' in \emph{Proceedings of the 2024 Conference on
  Empirical Methods in Natural Language Processing}, Y.~Al-Onaizan, M.~Bansal,
  and Y.-N. Chen, Eds.\hskip 1em plus 0.5em minus 0.4em\relax Miami, Florida,
  USA: Association for Computational Linguistics, Nov. 2024, pp.
  17\,432--17\,445. [Online]. Available:
  \url{https://aclanthology.org/2024.emnlp-main.965/}
\BIBentrySTDinterwordspacing

\bibitem{brinkmann2024mechanistic}
J.~Brinkmann, A.~Sheshadri, V.~Levoso, P.~Swoboda, and C.~Bartelt, ``A
  mechanistic analysis of a transformer trained on a symbolic multi-step
  reasoning task,'' \emph{arXiv preprint arXiv:2402.11917}, 2024.

\bibitem{yang2025emergent}
Y.~Yang, D.~I. Campbell, K.~Huang, M.~Wang, J.~D. Cohen, and T.~W. Webb,
  ``Emergent symbolic mechanisms support abstract reasoning in large language
  models,'' in \emph{Proceedings of the 42nd International Conference on
  Machine Learning (ICML)}, 2025.

\bibitem{wang2026improvingsparseautoencoderdynamic}
\BIBentryALTinterwordspacing
D.~Wang, J.~Zhang, D.~Su, and H.~Huang, ``Improving sparse autoencoder with
  dynamic attention,'' 2026. [Online]. Available:
  \url{https://arxiv.org/abs/2604.14925}
\BIBentrySTDinterwordspacing

\bibitem{tolooshams2026mechanisticinterpretabilitysparseautoencoder}
\BIBentryALTinterwordspacing
B.~Tolooshams, A.~Shen, and A.~Anandkumar, ``Mechanistic interpretability with
  sparse autoencoder neural operators,'' 2026. [Online]. Available:
  \url{https://arxiv.org/abs/2509.03738}
\BIBentrySTDinterwordspacing

\bibitem{elhage2022toy}
N.~Elhage, T.~Hume, C.~Olsson, N.~Schiefer, T.~Henighan, S.~Kravec,
  Z.~Hatfield-Dodds, R.~Lasenby, D.~Drain, C.~Chen \emph{et~al.}, ``Toy models
  of superposition,'' \emph{arXiv preprint arXiv:2209.10652}, 2022.

\bibitem{bricken2023monosemantic}
\BIBentryALTinterwordspacing
T.~Bricken, A.~Templeton, J.~Batson, B.~Chen, A.~Jermyn, T.~Conerly, N.~Turner,
  C.~Olsson, S.~Douglas, B.~McKenzie-Nelson, N.~Elhage, S.~Hume, J.~Knight,
  B.~Shlegeris, N.~Ninan, A.~Schlatter, C.~Olah, and T.~Henighan, ``Towards
  monosemanticity: Decomposing language models with dictionary learning,''
  \emph{Transformer Circuits Thread}, 2023, anthropic Research. [Online].
  Available:
  \url{https://transformer-circuits.pub/2023/monosemantic-features/index.html}
\BIBentrySTDinterwordspacing

\bibitem{gurnee2023finding}
W.~Gurnee, N.~Nanda, M.~Pauly, K.~Harvey, D.~Troitskii, and D.~Bertsimas,
  ``Finding neurons in a haystack: Case studies with sparse probing,''
  \emph{arXiv preprint arXiv:2305.01610}, 2023.

\bibitem{templeton2024scaling}
\BIBentryALTinterwordspacing
A.~Templeton, T.~Conerly, J.~Marcus, J.~Lindsay, T.~Bricken, B.~Chen,
  A.~Pearce, N.~Ley, N.~Russell, T.~Henighan, and C.~Olah, ``Scaling
  monosemanticity: Extracting interpretable features from claude 3 sonnet,''
  \emph{Transformer Circuits Thread}, 2024, anthropic. [Online]. Available:
  \url{https://transformer-circuits.pub/2024/scaling-monosemanticity/}
\BIBentrySTDinterwordspacing

\bibitem{lieberum2024gemmascopeopensparse}
\BIBentryALTinterwordspacing
T.~Lieberum, S.~Rajamanoharan, A.~Conmy, L.~Smith, N.~Sonnerat, V.~Varma,
  J.~Kram\'{a}r, A.~Dragan, R.~Shah, and N.~Nanda, ``Gemma scope: Open sparse
  autoencoders everywhere all at once on gemma 2,'' 2024. [Online]. Available:
  \url{https://arxiv.org/abs/2408.05147}
\BIBentrySTDinterwordspacing

\bibitem{shu2025surveysparseautoencodersinterpreting}
\BIBentryALTinterwordspacing
D.~Shu, X.~Wu, H.~Zhao, D.~Rai, Z.~Yao, N.~Liu, and M.~Du, ``A survey on sparse
  autoencoders: Interpreting the internal mechanisms of large language
  models,'' 2025. [Online]. Available: \url{https://arxiv.org/abs/2503.05613}
\BIBentrySTDinterwordspacing

\bibitem{rajamanoharan2024improvingdictionarylearninggated}
\BIBentryALTinterwordspacing
S.~Rajamanoharan, A.~Conmy, L.~Smith, T.~Lieberum, V.~Varma, J.~Kram\'{a}r,
  R.~Shah, and N.~Nanda, ``Improving dictionary learning with gated sparse
  autoencoders,'' 2024. [Online]. Available:
  \url{https://arxiv.org/abs/2404.16014}
\BIBentrySTDinterwordspacing

\bibitem{gao2024scalingevaluatingsparseautoencoders}
\BIBentryALTinterwordspacing
L.~Gao, T.~D. la~Tour, H.~Tillman, G.~Goh, R.~Troll, A.~Radford, I.~Sutskever,
  J.~Leike, and J.~Wu, ``Scaling and evaluating sparse autoencoders,'' 2024.
  [Online]. Available: \url{https://arxiv.org/abs/2406.04093}
\BIBentrySTDinterwordspacing

\bibitem{bussmann2024batchtopksparseautoencoders}
\BIBentryALTinterwordspacing
B.~Bussmann, P.~Leask, and N.~Nanda, ``Batchtopk sparse autoencoders,'' 2024.
  [Online]. Available: \url{https://arxiv.org/abs/2412.06410}
\BIBentrySTDinterwordspacing

\bibitem{ali2025entropy}
R.~Ali, F.~Caso, C.~Irwin, and P.~Li{\`o}, ``Entropy-lens: The information
  signature of transformer computations,'' \emph{arXiv preprint
  arXiv:2502.16570}, 2025.

\bibitem{karvonen2025activation}
\BIBentryALTinterwordspacing
A.~Karvonen, J.~Chua, C.~Dumas, K.~Fraser-Taliente, S.~Kantamneni, J.~Minder,
  E.~Ong, A.~S. Sharma, D.~Wen, O.~Evans, and S.~Marks, ``Activation oracles:
  Training and evaluating llms as general-purpose activation explainers,''
  Anthropic Alignment Science Blog, 2025, anthropic. [Online]. Available:
  \url{https://alignment.anthropic.com/2025/activation-oracles/}
\BIBentrySTDinterwordspacing

\bibitem{swann2026sparseautoencodersrevealinterpretable}
\BIBentryALTinterwordspacing
A.~Swann, L.~McGranahan, H.~Buurmeijer, M.~K. III, and M.~Schwager, ``Sparse
  autoencoders reveal interpretable and steerable features in vla models,''
  2026. [Online]. Available: \url{https://arxiv.org/abs/2603.19183}
\BIBentrySTDinterwordspacing

\bibitem{marks2025sparsefeaturecircuitsdiscovering}
\BIBentryALTinterwordspacing
S.~Marks, C.~Rager, E.~J. Michaud, Y.~Belinkov, D.~Bau, and A.~Mueller,
  ``Sparse feature circuits: Discovering and editing interpretable causal
  graphs in language models,'' 2025. [Online]. Available:
  \url{https://arxiv.org/abs/2403.19647}
\BIBentrySTDinterwordspacing

\bibitem{mcdougall2025faithfulness}
\BIBentryALTinterwordspacing
C.~McDougall, T.~Conerly, A.~Templeton, J.~Lindsey, T.~Bricken, B.~Chen,
  A.~Pearce, C.~Citro, E.~Ameisen, A.~Jones, H.~Cunningham, N.~L. Turner,
  M.~MacDiarmid, A.~Tamkin, E.~Durmus, T.~Hume, F.~Mosconi, C.~D. Freeman,
  T.~R. Sumers, E.~Rees, J.~Batson, A.~Jermyn, S.~Carter, C.~Olah, and
  T.~Henighan, ``Evaluating explanation faithfulness in toy models,''
  \emph{Transformer Circuits Thread}, 2025, anthropic. [Online]. Available:
  \url{https://transformer-circuits.pub/2025/faithfulness-toy-model/index.html}
\BIBentrySTDinterwordspacing

\bibitem{dunefsky2024transcodersinterpretablellmfeature}
\BIBentryALTinterwordspacing
J.~Dunefsky, P.~Chlenski, and N.~Nanda, ``Transcoders find interpretable llm
  feature circuits,'' 2024. [Online]. Available:
  \url{https://arxiv.org/abs/2406.11944}
\BIBentrySTDinterwordspacing

\bibitem{golimblevskaia2026circuitinsightsinterpretabilityactivations}
\BIBentryALTinterwordspacing
E.~Golimblevskaia, A.~Jain, B.~Puri, A.~Ibrahim, W.~Samek, and S.~Lapuschkin,
  ``Circuit insights: Towards interpretability beyond activations,'' 2026.
  [Online]. Available: \url{https://arxiv.org/abs/2510.14936}
\BIBentrySTDinterwordspacing

\bibitem{sinii2026smallvectorsbigeffects}
\BIBentryALTinterwordspacing
V.~Sinii, N.~Balagansky, G.~Gerasimov, D.~Laptev, Y.~Aksenov, V.~Kurochkin,
  A.~Gorbatovski, B.~Shaposhnikov, and D.~Gavrilov, ``Small vectors, big
  effects: A mechanistic study of rl-induced reasoning via steering vectors,''
  2026. [Online]. Available: \url{https://arxiv.org/abs/2509.06608}
\BIBentrySTDinterwordspacing

\bibitem{kang2025modelwhispersteeringvectors}
\BIBentryALTinterwordspacing
X.~Kang, D.~Shi, and L.~Chen, ``Model whisper: Steering vectors unlock large
  language models' potential in test-time,'' 2025. [Online]. Available:
  \url{https://arxiv.org/abs/2512.04748}
\BIBentrySTDinterwordspacing

\bibitem{nguyen2025flexibleframeworklinearrepresentation}
\BIBentryALTinterwordspacing
T.~Nguyen and Y.~Leng, ``Toward a flexible framework for linear representation
  hypothesis using maximum likelihood estimation,'' 2025. [Online]. Available:
  \url{https://arxiv.org/abs/2502.16385}
\BIBentrySTDinterwordspacing

\bibitem{bartoszcze2025representationengineeringlargelanguagemodels}
\BIBentryALTinterwordspacing
L.~Bartoszcze, S.~Munshi, B.~Sukidi, J.~Yen, Z.~Yang, D.~Williams-King, L.~Le,
  K.~Asuzu, and C.~Maple, ``Representation engineering for large-language
  models: Survey and research challenges,'' 2025. [Online]. Available:
  \url{https://arxiv.org/abs/2502.17601}
\BIBentrySTDinterwordspacing

\bibitem{turner2024steeringlanguagemodelsactivation}
\BIBentryALTinterwordspacing
A.~M. Turner, L.~Thiergart, G.~Leech, D.~Udell, J.~J. Vazquez, U.~Mini, and
  M.~MacDiarmid, ``Steering language models with activation engineering,''
  2024. [Online]. Available: \url{https://arxiv.org/abs/2308.10248}
\BIBentrySTDinterwordspacing

\bibitem{panickssery2024steeringllama2contrastive}
\BIBentryALTinterwordspacing
N.~Panickssery, N.~Gabrieli, J.~Schulz, M.~Tong, E.~Hubinger, and A.~M. Turner,
  ``Steering llama 2 via contrastive activation addition,'' 2024. [Online].
  Available: \url{https://arxiv.org/abs/2312.06681}
\BIBentrySTDinterwordspacing

\bibitem{sofroniew2026emotion}
\BIBentryALTinterwordspacing
N.~Sofroniew, J.~Lindsey, W.~Gurnee, E.~Ameisen, B.~Chen, A.~Pearce, N.~L.
  Turner, C.~Citro, D.~Abrahams, S.~Carter, B.~Hosmer, J.~Marcus, M.~Sklar,
  A.~Templeton, T.~Bricken, C.~McDougall, H.~Cunningham, T.~Henighan,
  A.~Jermyn, A.~Jones, A.~Persic, Z.~Qi, T.~B. Thompson, S.~Zimmerman,
  K.~Rivoire, T.~Conerly, C.~Olah, and J.~Batson, ``Emotion concepts and their
  function in a large language model,'' \emph{Transformer Circuits Thread},
  2026, anthropic. [Online]. Available:
  \url{https://transformer-circuits.pub/2026/emotions/index.html}
\BIBentrySTDinterwordspacing

\bibitem{soo2025interpretablesteeringlargelanguage}
\BIBentryALTinterwordspacing
S.~Soo, C.~Guang, W.~Teng, C.~Balaganesh, T.~Guoxian, and Y.~Ming,
  ``Interpretable steering of large language models with feature guided
  activation additions,'' 2025. [Online]. Available:
  \url{https://arxiv.org/abs/2501.09929}
\BIBentrySTDinterwordspacing

\bibitem{wedgwood2026dspadynamicsaesteering}
\BIBentryALTinterwordspacing
J.~Wedgwood, A.~Muhamed, M.~T. Diab, and V.~Smith, ``Dspa: Dynamic sae steering
  for data-efficient preference alignment,'' 2026. [Online]. Available:
  \url{https://arxiv.org/abs/2603.21461}
\BIBentrySTDinterwordspacing

\bibitem{wang2023foldseefficientrulebasedmachine}
\BIBentryALTinterwordspacing
H.~Wang and G.~Gupta, ``Fold-se: An efficient rule-based machine learning
  algorithm with scalable explainability,'' 2023. [Online]. Available:
  \url{https://arxiv.org/abs/2208.07912}
\BIBentrySTDinterwordspacing

\bibitem{padalkar2023nesyfoldneurosymbolicframeworkinterpretable}
\BIBentryALTinterwordspacing
P.~Padalkar, H.~Wang, and G.~Gupta, ``Nesyfold: Neurosymbolic framework for
  interpretable image classification,'' 2023. [Online]. Available:
  \url{https://arxiv.org/abs/2301.12667}
\BIBentrySTDinterwordspacing

\bibitem{PADALKAR_2025}
\BIBentryALTinterwordspacing
P.~Padalkar and G.~Gupta, ``Symbolic rule extraction from attention-guided
  sparse representations in vision transformers,'' \emph{Theory and Practice of
  Logic Programming}, vol.~25, no.~4, pp. 722--738, Jul. 2025. [Online].
  Available: \url{http://dx.doi.org/10.1017/S1471068425100318}
\BIBentrySTDinterwordspacing

\bibitem{pan2023logiclm}
L.~Pan, A.~Albalak, X.~Wang, and W.~Y. Wang, ``Logic-lm: Empowering large
  language models with symbolic solvers for faithful logical reasoning,'' in
  \emph{Findings of the Association for Computational Linguistics: EMNLP 2023},
  2023.

\bibitem{kambhampati2024modulo}
S.~Kambhampati, K.~Valmeekam, L.~Guan, M.~Verma, K.~Stechly, S.~Bhambri,
  L.~Saldyt, and A.~Murthy, ``Llms can't plan, but can help planning in
  llm-modulo frameworks,'' in \emph{Proceedings of the 41st International
  Conference on Machine Learning (ICML)}, 2024.

\bibitem{makelov2024principledevaluationssparseautoencoders}
\BIBentryALTinterwordspacing
A.~Makelov, G.~Lange, and N.~Nanda, ``Towards principled evaluations of sparse
  autoencoders for interpretability and control,'' 2024. [Online]. Available:
  \url{https://arxiv.org/abs/2405.08366}
\BIBentrySTDinterwordspacing

\end{thebibliography}
}
\end{document}